\documentclass[10pt,twocolumn,letterpaper]{article}
\usepackage[pagenumbers]{cvpr}

\usepackage{microtype}

\definecolor{cvprblue}{rgb}{0.21,0.49,0.74}
\usepackage[pagebackref,breaklinks,colorlinks,allcolors=cvprblue]{hyperref}
\usepackage{colortbl}
\usepackage{xcolor}
\usepackage{multirow}

\title{Learning to Act Robustly with View-Invariant Latent Actions}

\author{
Youngjoon Jeong$^{1}$\thanks{There authors contributed equally},
Junha Chun$^{2}$\footnotemark[1],
Taesup Kim$^{1}$\thanks{Corresponding author. Email: \texttt{taesup.kim@snu.ac.kr}}\\[2mm]
$^1$Graduate School of Data Science, Seoul National University \\
$^2$Department of Electrical and Computer Engineering, Seoul National University
}
\begin{document}

\twocolumn[{
    \maketitle
    \begin{center}
    \includegraphics[width=0.90\linewidth]{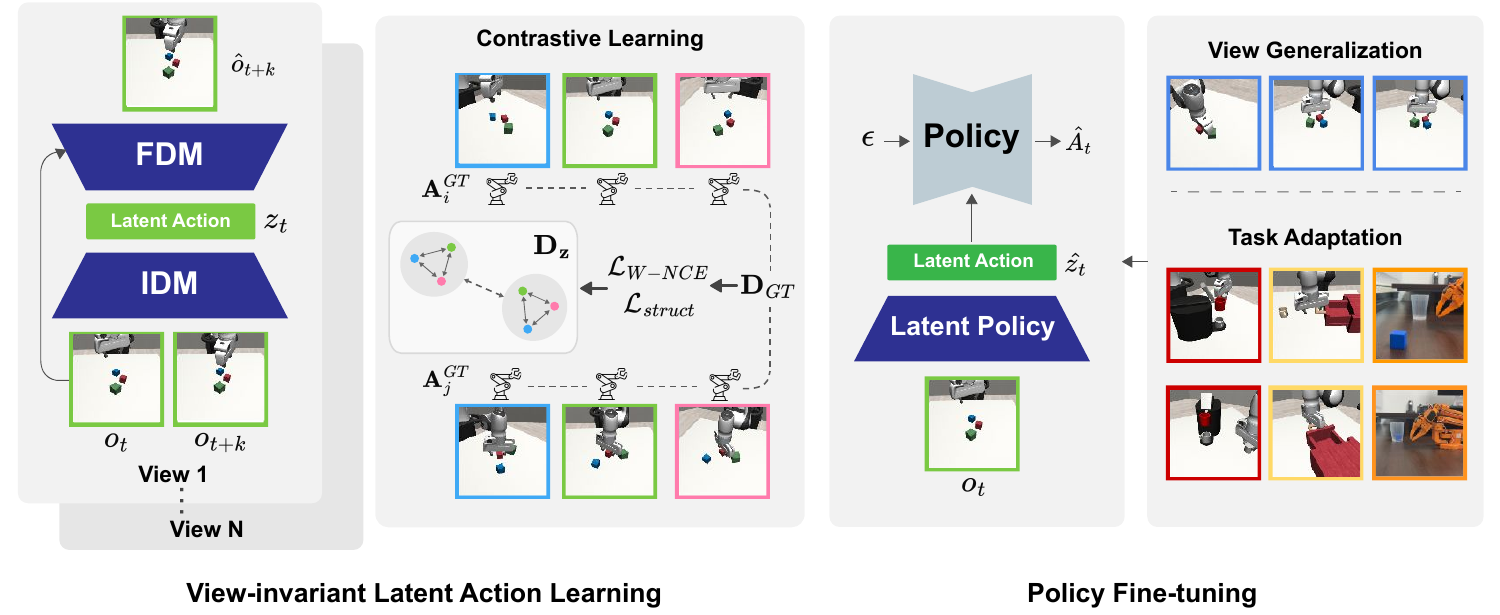}
    \end{center}
    \vspace{-0.5cm}
    \captionsetup{type=figure}
    \captionof{figure}{%
    \textbf{VILA Overview.} Our method learns view-invariant latent actions by aligning them using action-aware contrastive learning along with predicting the future. A latent policy that predicts these latent actions from the current observation is then used as a vision encoder to condition a downstream visuomotor policy, yielding robust view generalization and task adaptation in simulation and the real-world.
    }\label{fig:teaser}
    \vspace{0.3cm}
}]

{
  \renewcommand{\thefootnote}{\fnsymbol{footnote}} 
  \footnotetext[1]{These authors contributed equally}
  \footnotetext[2]{Corresponding author. Email: \texttt{taesup.kim@snu.ac.kr}}
}

\begin{abstract}
Vision-based robotic policies often struggle with even minor viewpoint changes, underscoring the need for view-invariant visual representations. This challenge becomes more pronounced in real-world settings, where viewpoint variability is unavoidable and can significantly disrupt policy performance.
Existing methods typically learn invariance from multi-view observations at the scene level, but such approaches rely on visual appearance and fail to incorporate the physical dynamics essential for robust generalization.
We propose View-Invariant Latent Action (VILA), which models a latent action capturing transition patterns across trajectories to learn view-invariant representations grounded in physical dynamics. VILA aligns these latent actions across viewpoints using an action-guided objective based on ground-truth action sequences.
Experiments in both simulation and the real world show that VILA-based policies generalize effectively to unseen viewpoints and transfer well to new tasks, establishing VILA as a strong pretraining framework that improves robustness and downstream learning performance. Demonstration videos can be found on our website: \url{https://joon-stack.github.io/VILA/}.
\end{abstract}      
\section{Introduction}
\label{sec:intro}

Vision-based robotic policies are brittle to changes in camera viewpoint, posing a critical barrier to robust real-world deployment. While collecting large-scale, multi-view datasets is a straightforward way to expose policies to viewpoint variation, it incurs prohibitive data acquisition costs and scales poorly.

Prior approaches have largely focused on obtaining a \emph{scene-level} visual representation that is stable under viewpoint changes.
This is often achieved either by modifying observations (e.g., via novel view synthesis or geometric inputs)~\citep{jiang2025knowcameraisviewinvariant, tian2025viewinvariantpolicylearningzeroshot, chen2024roviaugrobotviewpointaugmentation} or by pre-training encoders to learn invariant features~\citep{pang2025reviwo, lee2025class, seo2023multiviewmaskedworldmodels}.
Despite these different strategies, the underlying goal remains the same: enforcing robustness at the image-level.
In practice, this means that a single compact feature vector is asked to summarize the entire image, including static layout and background, and at the same time carry all information about task-relevant motion.
Enforcing viewpoint robustness at this level can therefore be unnecessarily demanding: it asks for invariance over a representation that is much broader than what is actually needed for control, and it does not explicitly distinguish between static context and the underlying dynamics that drive actions.

In this paper, we propose a different, more targeted approach. 
\textbf{Our key insight is that invariance should be enforced not on a static \emph{scene-level visual representation}, but rather on the \emph{dynamics}, the \emph{change} in the scene that is relevant to the actions.}
Representations of change are naturally more compact and predominantly capture how the agent and objects move, rather than how the entire scene looks from a particular viewpoint. 
To this end, we introduce \textbf{V}iew-\textbf{I}nvariant \textbf{L}atent \textbf{A}ction (\textbf{VILA}), a novel pre-training framework described in Figure~\ref{fig:teaser}.
VILA builds on the notion of latent actions~\citep{schmidt2024learningactactions, ye2025latentactionpretrainingvideos, nikulin2025latentactionlearningrequires, bauer2025latentactiondiffusioncrossembodiment}, which model dynamics as a compact code explaining the observed change between consecutive observations, but with a critical modification.
VILA learns a latent action representation that encodes the change between observations. 
At the same time, we use ground-truth (GT) action sequences as \emph{action-guided} alignment to shape this space with a weighted contrastive loss and a global similarity-alignment term, enforcing view-invariance directly in this dynamics-centered latent space.

\section{Related Works}
\label{sec:related_works}

\subsection{Viewpoint Robustness in Visuomotor Policies}
\label{sec:viewpoint-robustness}

A key challenge in visuomotor policy learning is generalizing to camera poses that differ between training and deployment. Prior work mainly tackles this via modifying the observations or learning view-robust representations.

The first family operates at the \emph{observation level}. Multi-view calibrated datasets~\citep{fang2024rh20t, walke2023bridgedata, DROID, RoboHive} expose policies to multiple cameras, but typically with limited viewpoint diversity. Data augmentation methods instead synthesize or collect additional views so that the policy sees a wider range of poses; several recent approaches~\citep{tian2025viewinvariantpolicylearningzeroshot, chen2024roviaugrobotviewpointaugmentation, ding2025imagination} use novel view synthesis (NVS) to expand labeled datasets. Other methods explicitly condition the policy on camera geometry, as in ~\citep{jiang2025knowcameraisviewinvariant}, which provides intrinsics, extrinsics, or per-pixel ray embeddings as additional inputs.

The second family operates at the \emph{representation level} by learning view-robust visual encoders~\citep{seo2023multiviewmaskedworldmodels, pang2025reviwo, lee2025class, Shi2025NVSPolicyAN, li2024RoboUniView}. These methods train scene-level features to remain stable under camera motion while summarizing the entire image, which can conflate task-relevant motion with static context or visual appearances.

In contrast, VILA does not impose invariance on a scene-level representation. We enforce invariance only on the latent action, a compact representation of the system's dynamics, so that model capacity is focused on how the agent and objects move rather than on full-scene appearance.

\subsection{Latent Actions}

Latent action models~\citep{schmidt2024learningactactions, nikulin2025latentactionlearningrequires, bauer2025latentactiondiffusioncrossembodiment} learn compact dynamics representations by encoding the change between two observations $o_t$ and $o_{t+k}$ into a latent action $z$. 
This is typically done with an inverse dynamics model (IDM) inferring $z = \mathrm{IDM}(o_t, o_{t+k})$ and a forward dynamics model (FDM) reconstructing $\hat{o}_{t+k} = \mathrm{FDM}(o_t, z)$. 
To avoid trivial solutions, $z$ is constrained by a low-dimensional bottleneck or vector quantization~\citep{oord2018neuraldiscreterepresentationlearning,schmidt2024learningactactions, ye2025latentactionpretrainingvideos}.

These dynamics representations have proved useful priors for world models~\citep{bruce2024geniegenerativeinteractiveenvironments, gao2025adaworld, ren2025videoworldexploringknowledgelearning} and policies~\citep{schmidt2024learningactactions, ye2025latentactionpretrainingvideos, nikulin2025latentactionlearningrequires, bu2025univla, agibotworldcontributors2025agibotworldcolosseolargescale, chen2025villaxenhancinglatentaction,bauer2025latentactiondiffusioncrossembodiment}. 
However, existing work mainly optimizes latent actions to be predictive and useful for control, without explicitly targeting robustness to camera pose changes. 
Because latent actions are defined over changes between observations, they already emphasize motion rather than static appearance, which makes them a natural place to enforce additional viewpoint invariance. 
In VILA, we keep the standard latent action learning objective but add multi-view, action-guided regularization that forces latent actions for the same underlying motion to align across viewpoints, and then use this space as the interface for learning viewpoint-robust visuomotor policies.

\section{Methods}
\label{sec:methods}

Our proposed framework consists of two stages:
(i) \emph{latent action learning}, where we learn a compact, action-guided and view-invariant dynamics representation, and
(ii) \emph{latent behavior cloning}, where we train a latent policy that takes the current observation as input and predicts latent actions, so that this latent policy serves as a view-robust vision encoder that conditions a downstream visuomotor policy.

In the latent action learning stage, we pursue two primary objectives:
(i) learning a base latent action representation that captures the underlying dynamics in a compact latent space; and
(ii) enforcing view-invariance on this latent action using an action-aware contrastive objective.
We build our base latent action learner on top of the LAOM framework~\citep{schmidt2024learningactactions}, and then introduce our action-aware contrastive loss and structural regularizer.
The resulting IDM is then reused in the latent behavior cloning stage, where we train a latent policy as an encoder.

\subsection{Base Latent Action Learning}
\label{sec:base_latent_action}

We first learn a base latent action representation following the core design of LAOM~\citep{schmidt2024learningactactions}, which relies on a temporal consistency loss $\mathcal{L}_{\text{LA}}$ in a compact latent space.

We index time by subscripts $t$ and camera viewpoints by superscripts $v$, so $o_t^v$ denotes the observation at time $t$ from view $v$. 
A visual encoder $E$ maps each observation $o_t^v$ to a feature $s_t^v = E(o_t^v)$, and we sample a temporal offset $k \in \{1,\dots,K\}$ to obtain $s_{t+k}^v = E(o_{t+k}^v)$. 
The IDM infers a latent action $z_t^v = \mathrm{IDM}(s_t^v, s_{t+k}^v)$, and the FDM predicts the future feature $\hat{s}_{t+k}^v = \mathrm{FDM}(s_t^v, z_t^v)$.

The core temporal consistency loss $\mathcal{L}_{\text{LA}}$ is defined as the mean squared error between this prediction $\hat{s}_{t+k}^v$ and a stable target $s^{\text{tgt},v}_{t+k}$. 
The target is obtained by passing $o_{t+k}^v$ through a separate, non-backpropagated encoder $E^{\text{tgt}}$, whose parameters are an EMA of the online encoder $E$. 
Let $\mathcal{D}_k$ denote the set of all pairs $(o_t^v, o_{t+k}^v)$ for which the segment $(t, t+k)$ exists in the dataset. 
Our latent action loss is
\begin{equation}
    \mathcal{L}_{\text{LA}}
    =
    \mathbb{E}_{k \sim \mathcal{U}(1,K),\, (o_t^v, o_{t+k}^v) \sim \mathcal{D}_k}
    \bigl[
        \| \mathrm{FDM}(s_t^v, z_t^v) - s^{\text{tgt},v}_{t+k} \|_2^2
    \bigr],
    \label{eq:latent_action_loss}
\end{equation}
where $z_t^v = \mathrm{IDM}(s_t^v, s_{t+k}^v)$.
This loss encourages $(E,\mathrm{IDM},\mathrm{FDM})$ to discover a compact latent action $z_t^v$ that explains the change from $o_t^v$ to $o_{t+k}^v$ without reconstructing pixels.
We use the same mini-batch construction described in Sec.~\ref{sec:action_contrastive} to obtain training samples for this objective.

\begin{table*}[t]
    \centering
    \caption{\textbf{Unseen View Generalization in Fine-tuned and Frozen Settings (Simulation).} For each of the 25 views in Figure~\ref{fig:25views}, we evaluate average success rates (\%) over 20 episodes and report for 10 seen views, 15 unseen views, and their ratio denoted as Rel.}
    \label{tab:view_generalization_combined}
    \resizebox{\textwidth}{!}{
    \begin{tabular}{l *{5}{rrr}}
        \toprule
        \multirow{2}{*}{\textbf{Model}}& \multicolumn{3}{c}{\textbf{Lift}} & \multicolumn{3}{c}{\textbf{Square}} & \multicolumn{3}{c}{\textbf{Stack Three}} & \multicolumn{3}{c}{\textbf{Coffee}} & \multicolumn{3}{c}{\textbf{Mug Cleanup}} \\
        \cmidrule(lr){2-4} \cmidrule(lr){5-7} \cmidrule(lr){8-10} \cmidrule(lr){11-13} \cmidrule(lr){14-16}
         & Seen & Unseen & Rel. & Seen & Unseen & Rel. & Seen & Unseen & Rel. & Seen & Unseen & Rel. & Seen & Unseen & Rel. \\
        \midrule\midrule
        \multicolumn{16}{l}{\textbf{Fine-Tuned}} \\
        \midrule
        \rowcolor{blue!20}
        VILA (Ours) & 99.50 & \textbf{94.70} & \textbf{95.18} & 69.00 & \textbf{19.80} & \textbf{28.70} & 69.00 & \textbf{53.65} & \textbf{77.75} & 63.00 & \textbf{12.65} & \textbf{20.08} & 56.75 & \textbf{27.85} & \textbf{49.07} \\
        Vanilla & 97.00 & 77.00 & 79.38 & 57.00 & 8.70 & 15.26 & 64.00 & 23.70 & 37.03 & 63.00 & 0.35 & 0.56 & 55.50 & 9.70 & 17.48 \\
        CLASS & 96.50 & 65.00 & 67.36 & 54.00 & 9.00 & 16.67 & 30.00 & 10.35 & 34.50 & 28.00 & 0.35 & 1.25 & 22.00 & 6.35 & 28.86 \\
        ReViWo & 83.00 & 38.00 & 45.78 & 5.00 & 0.35 & 7.00 & 0.00 & 0.00 & 0.00 & 0.00 & 0.00 & 0.00 & 0.00 & 0.00 & 0.00 \\
        KYC & 72.00 & 46.00 & 63.89 & 14.00 & 0.00 & 0.00 & 1.90 & 0.00 & 0.00 & 10.70 & 0.00 & 0.00 & 4.00 & 0.00 & 0.00 \\
        \midrule\midrule
        \multicolumn{16}{l}{\textbf{Frozen}} \\ 
        \midrule
        \rowcolor{blue!20}
        VILA (Ours) & 72.00 & 60.70 & \textbf{84.31} & 34.50 & \textbf{10.35} & \textbf{30.00} & 15.50 & \textbf{6.65} & \textbf{42.90} & 29.50 & \textbf{4.00} & \textbf{13.56} & 33.50 & \textbf{15.00} & \textbf{44.78} \\
        Vanilla & 76.00 & 47.30 & 62.24 & 14.00 & 0.35 & 2.50 & 0.50 & 0.00 & 0.00 & 6.50 & 0.00 & 0.00 & 5.00 & 0.00 & 0.00 \\
        CLASS & 93.90 & \textbf{65.00} & 69.22 & 25.00 & 0.35 & 1.40 & 0.50 & 0.00 & 0.00 & 6.00 & 0.00 & 0.00 & 4.00 & 0.00 & 0.00 \\
        ReViWo & 78.00 & 40.70 & 52.18 & 1.50 & 0.00 & 0.00 & 0.00 & 0.00 & 0.00 & 0.50 & 0.00 & 0.00 & 0.00 & 0.00 & 0.00 \\
        \bottomrule
    \end{tabular}
    }
\end{table*}

\subsection{Action-Guided Latent Action Invariance}
\label{sec:action_contrastive}

To make the latent action invariant to camera viewpoint, we introduce an action-aware contrastive objective. 
The key idea is that latent actions inferred from different viewpoints should be close whenever their corresponding future GT action sequences are similar.

We construct each training batch as follows. 
We first sample a temporal offset $k \in \{1,\dots,K\}$ and then sample $N$ base time indices $\{t_i\}_{i=1}^N$ such that the segment $(t_i, t_i + k)$ and its GT action sequence $\mathbf{A}_i^{\text{GT}} = (a_{t_i}, \dots, a_{t_i+k-1})$ are available. 
For each base index $t_i$, we sample $V$ random camera viewpoints and retrieve the corresponding observation pairs $\{(o_{t_i}^v, o_{t_i+k}^v)\}_{v=1}^V$, yielding $B = N V$ latent action samples per batch. 
We index these $B$ transitions by a single index $i \in \{1,\dots,B\}$, so each $i$ implicitly corresponds to a particular $(t, v)$ pair, and denote the associated latent action by $z_i$ and GT action sequence by $\mathbf{A}_i^{\text{GT}}$. 
Within each batch, $k$ is held fixed so that all samples share the same prediction horizon.

We use the GT action sequences as supervision. 
For each sample $i$, let $\mathbf{A}_i^{\text{GT}} \in \mathbb{R}^{k \times D}$ denote the sequence of $k$ actions. 
We first define a normalized squared distance between two sequences
\begin{equation}
    d_{ij}
    =
    \frac{
        \|\mathbf{A}_i^{\text{GT}} - \mathbf{A}_j^{\text{GT}}\|_F^2
    }{
        kD
    },
    \label{eq:action_distance}
\end{equation}
where $\|\cdot\|_F$ is the Frobenius norm. 
We then convert these distances into soft weights
\begin{equation}
    w_{ij}
    =
    \frac{
        \exp(- d_{ij} / \beta)
    }{
        \sum_{\ell=1}^{B} \exp(- d_{i\ell} / \beta)
    },
    \label{eq:action_weights}
\end{equation}
where $\beta > 0$ controls the sharpness of the distribution; larger $w_{ij}$ indicate more similar action sequences.

Using these weights, we employ a weighted InfoNCE (supervised contrastive) loss~\citep{khosla2021supervisedcontrastivelearning, kim2025contrastiverepresentationregularizationvisionlanguageaction} that refines the \emph{local} structure of the latent action space:
\begin{equation}
\mathcal{L}_{\mathrm{W\text{-}NCE}}
= - \sum_{i=1}^{B} \sum_{j=1,j\neq i}^{B} 
w_{ij}\, 
\log 
\frac{
    \exp(\mathrm{sim}(z_i, z_j)/\tau)
}{
    \sum_{\ell=1}^{B} \exp(\mathrm{sim}(z_i, z_\ell)/\tau)
},
\label{eq:wnce}
\end{equation}
where $\mathrm{sim}(z_i, z_j) = \frac{z_i^\top z_j}{\|z_i\| \, \|z_j\|}$ is cosine similarity and $\tau>0$ is a temperature.

To capture the \emph{global} structure, we introduce an auxiliary structural loss based on pairwise similarities. 
We treat each GT action sequence $\mathbf{A}_i^{\text{GT}} \in \mathbb{R}^{k \times D}$ as a single vector in $\mathbb{R}^{kD}$ by flattening it (we reuse the same notation for brevity). 
We then L2-normalize latent actions and GT action vectors,
$\hat{z}_i = z_i / \| z_i \|_2$ and $\hat{\mathbf{A}}_i^{\text{GT}} = \mathbf{A}_i^{\text{GT}} / \| \mathbf{A}_i^{\text{GT}} \|_2$, and form cosine-similarity matrices
$S_z = \hat{Z}\hat{Z}^\top$ and $S_{\text{GT}} = \hat{A}\hat{A}^\top$, where the $i$-th rows of $\hat{Z}$ and $\hat{A}$ are $\hat{z}_i^\top$ and $(\hat{\mathbf{A}}_i^{\text{GT}})^\top$, respectively. 
We then align these global similarity structures via
\begin{equation}
    \mathcal{L}_{\text{struct}}
    =
    \| S_{\text{GT}} - S_z \|_F^2.
    \label{eq:struct}
\end{equation}

The total VILA representation loss is
\begin{equation}
    \mathcal{L}_{\text{VILA}}
    =
    \mathcal{L}_{\text{LA}}
    +
    \lambda_1 \mathcal{L}_{\text{W-NCE}}
    +
    \lambda_2 \mathcal{L}_{\text{struct}},
    \label{eq:vila_loss}
\end{equation}
where $\lambda_1$ and $\lambda_2$ are weighting hyperparameters.

\subsection{Latent Behavior Cloning}

The latent behavior cloning stage learns a policy that predicts latent actions from the current observation so that no future frames are needed at test time. 
We train a latent policy $\pi_{z}$ via behavior cloning:
\begin{equation}
    \mathcal{L}_{\mathrm{BC}}
    =
    \bigl\|
        \pi_{z}(s_t^v) - \mathrm{IDM}(s_t^v, s_{t+k}^v)
    \bigr\|_2^2.
    \label{eq:latent_bc}
\end{equation}

Since $\pi_{z}$ operates in the pre-trained latent action space, it inherits its view-invariant, structured properties. 
During fine-tuning, $\pi_{z}$ predicts latent actions from the current observation, and these are used as conditions for a downstream policy that outputs low-level actions.

\section{Experimental Results}
\label{sec:experiments}

\begin{figure}[t]
  \centering
  \begin{minipage}[c]{0.03\linewidth}
    \centering
    \rotatebox{90}{\scriptsize\textbf{Simulation (Robosuite)}}
  \end{minipage}
  \hfill
  \begin{minipage}[c]{0.95\linewidth}
    \begin{subfigure}[b]{0.32\linewidth}
      \includegraphics[width=\linewidth]{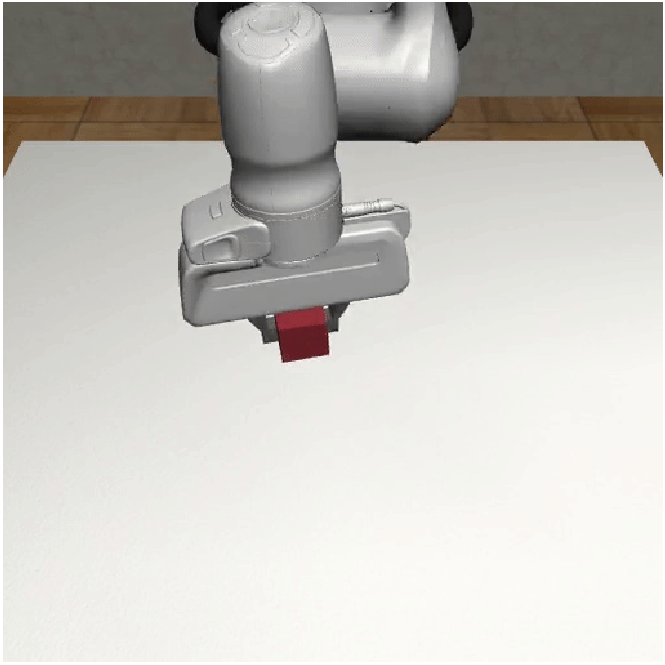}
      \caption{Lift}
      \label{fig:lift}
    \end{subfigure}
    \hfill
    \begin{subfigure}[b]{0.32\linewidth}
      \includegraphics[width=\linewidth]{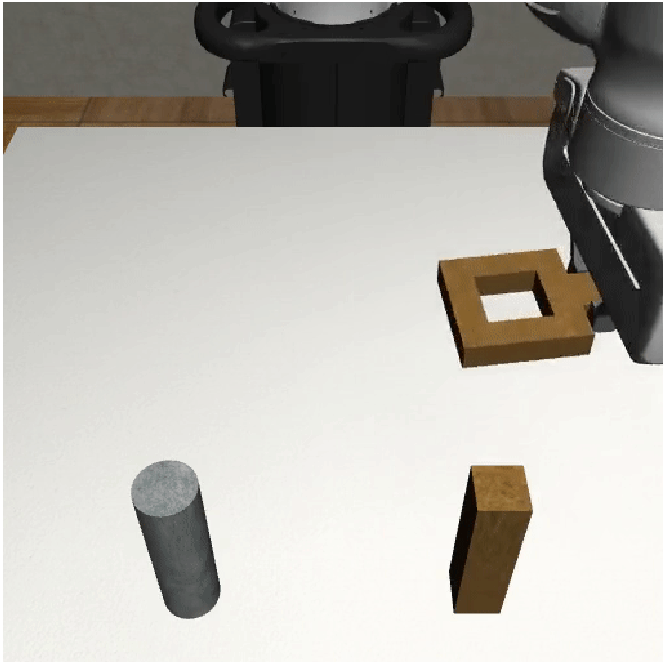}
      \caption{Square}
      \label{fig:square}
    \end{subfigure}
    \hfill
    \begin{subfigure}[b]{0.32\linewidth}
      \includegraphics[width=\linewidth]{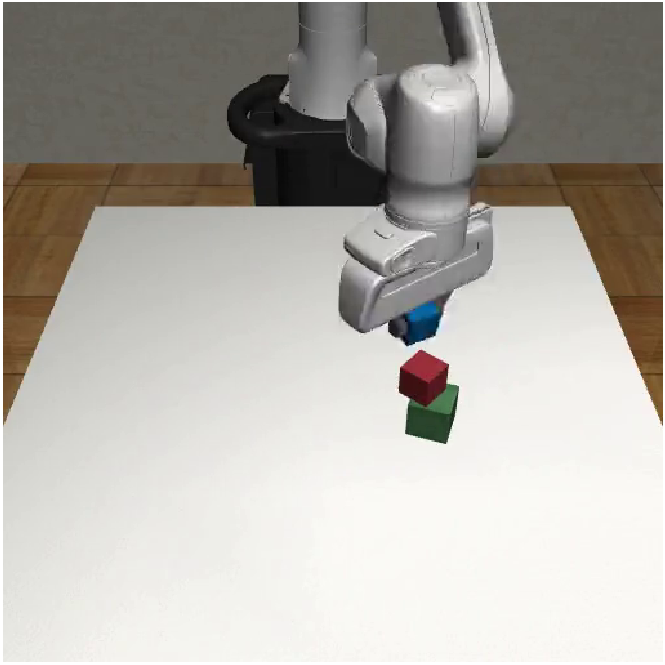}
      \caption{Stack Three}
      \label{fig:stackthree}
    \end{subfigure}

    \vspace{0.3em}

    \centering
    \begin{subfigure}[b]{0.32\linewidth}
      \includegraphics[width=\linewidth]{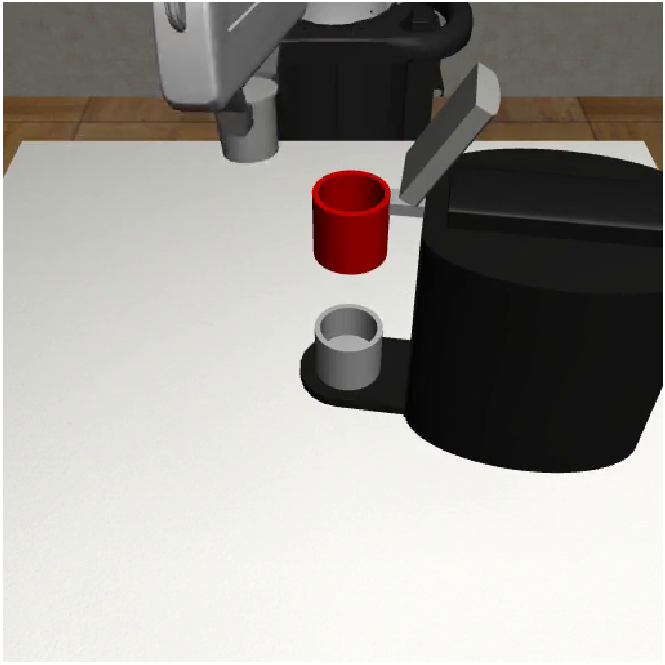}
      \caption{Coffee}
      \label{fig:coffee}
    \end{subfigure}%
    \hspace{0.02\linewidth}
    \begin{subfigure}[b]{0.32\linewidth}
      \includegraphics[width=\linewidth]{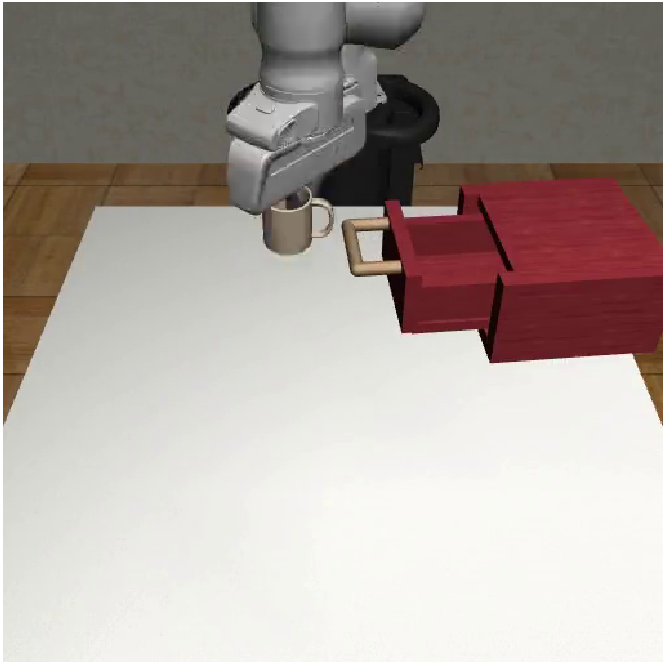}
      \caption{Mug Cleanup}
      \label{fig:mugcleanup}
    \end{subfigure}
  \end{minipage}

  \vspace{1em}
  \begin{minipage}[c]{0.03\linewidth}
    \centering
    \rotatebox{90}{\scriptsize\textbf{Real-World}}
  \end{minipage}
  \hfill
  \begin{minipage}[c]{0.95\linewidth}
    \centering
    \begin{subfigure}[b]{0.32\linewidth}
      \includegraphics[width=\linewidth]{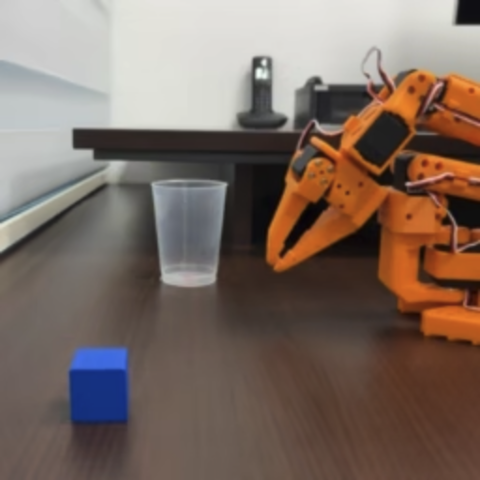}
      \caption{Pick \& Place}
      \label{subfig:pnp}
    \end{subfigure}%
    \hspace{0.02\linewidth}
    \begin{subfigure}[b]{0.32\linewidth}
      \includegraphics[width=\linewidth]{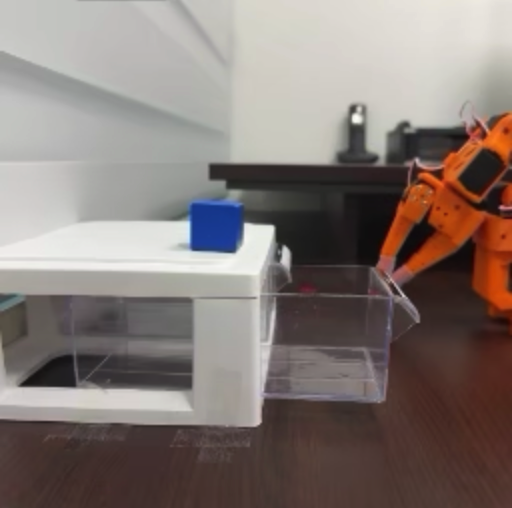}
      \caption{Drawer}
      \label{subfig:drawer}
    \end{subfigure}
  \end{minipage}

  \caption{
    \textbf{Dataset Overview.} All images are displayed at the same scale. Top rows show simulation tasks, and the bottom row shows real-world experiments.
  }
  \label{fig:dataset}
\end{figure}

\begin{figure}
  \centering
  
  \begin{minipage}[c]{0.03\columnwidth}
    \centering
    \rotatebox{90}{\small\textbf{Pick \& Place}}
  \end{minipage}
  \hfill
  \begin{minipage}[c]{0.95\columnwidth}
    \centering
    \begin{subfigure}[c]{0.32\linewidth}
      \includegraphics[width=\linewidth]{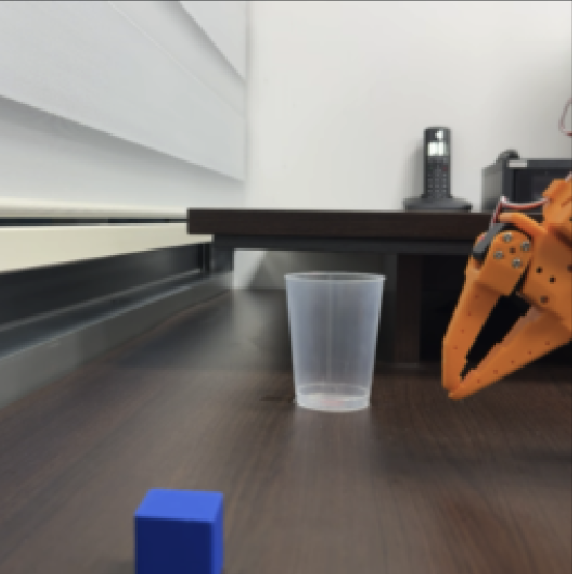}
      \caption{View 1}
      \label{fig:sub1}
    \end{subfigure}
    \hfill
    \begin{subfigure}[c]{0.32\linewidth}
      \includegraphics[width=\linewidth]{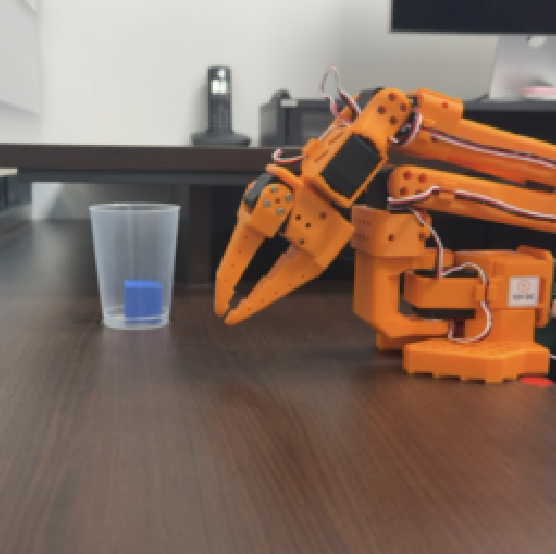}
      \caption{View 2}
      \label{fig:sub2}
    \end{subfigure}
    \hfill
    \begin{subfigure}[c]{0.32\linewidth}
      \includegraphics[width=\linewidth]{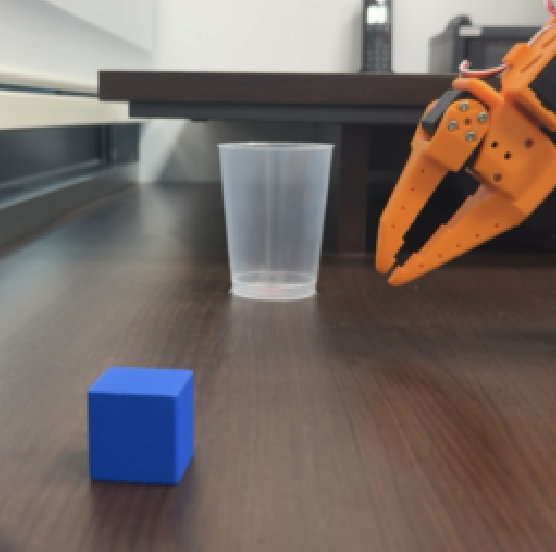}
      \caption{View 3}
      \label{fig:sub3}
    \end{subfigure}
  \end{minipage}

  \vspace{1em}

  \begin{minipage}[c]{0.03\columnwidth}
    \centering
    \rotatebox{90}{\small\textbf{Drawer}}
  \end{minipage}
  \hfill
  \begin{minipage}[c]{0.95\columnwidth}
    \centering
    \begin{subfigure}[c]{0.32\linewidth}
      \includegraphics[width=\linewidth]{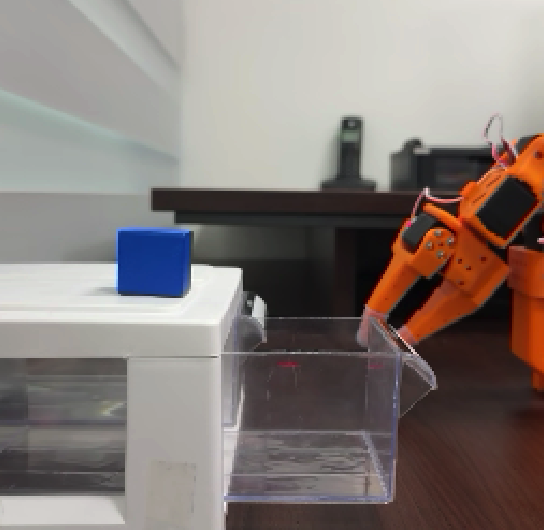}
      \caption{View 1}
      \label{fig:sub1_drawer}
    \end{subfigure}%
    \hspace{0.02\linewidth}
    \begin{subfigure}[c]{0.32\linewidth}
      \includegraphics[width=\linewidth]{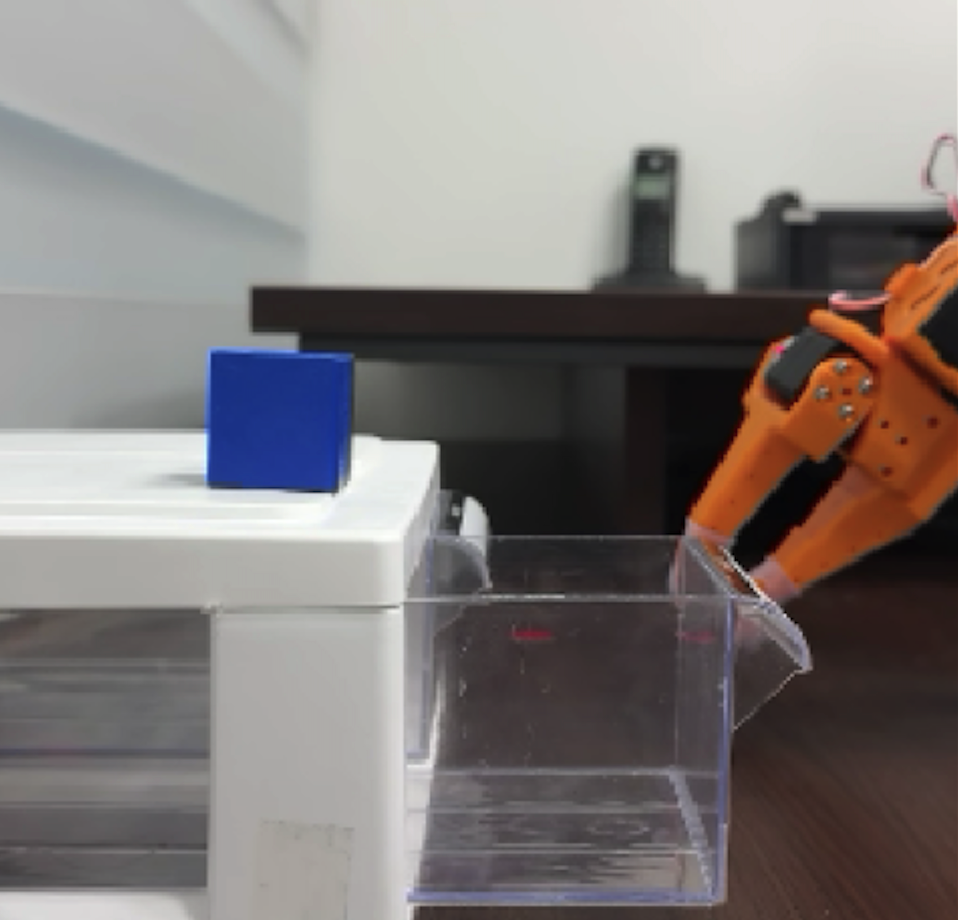}
      \caption{View 2}
      \label{fig:sub2_drawer}
    \end{subfigure}
  \end{minipage}

  \caption{
  \textbf{Real-world Unseen Views.} Rows show different tasks: (Top) \textbf{Pick \& Place}, (Bottom) \textbf{Drawer}.
  }
  \label{fig:real_world_views}
\end{figure}

\begin{figure}
  \centering
  \begin{subfigure}{0.86\columnwidth}
    \centering 
    \includegraphics[width=\columnwidth]{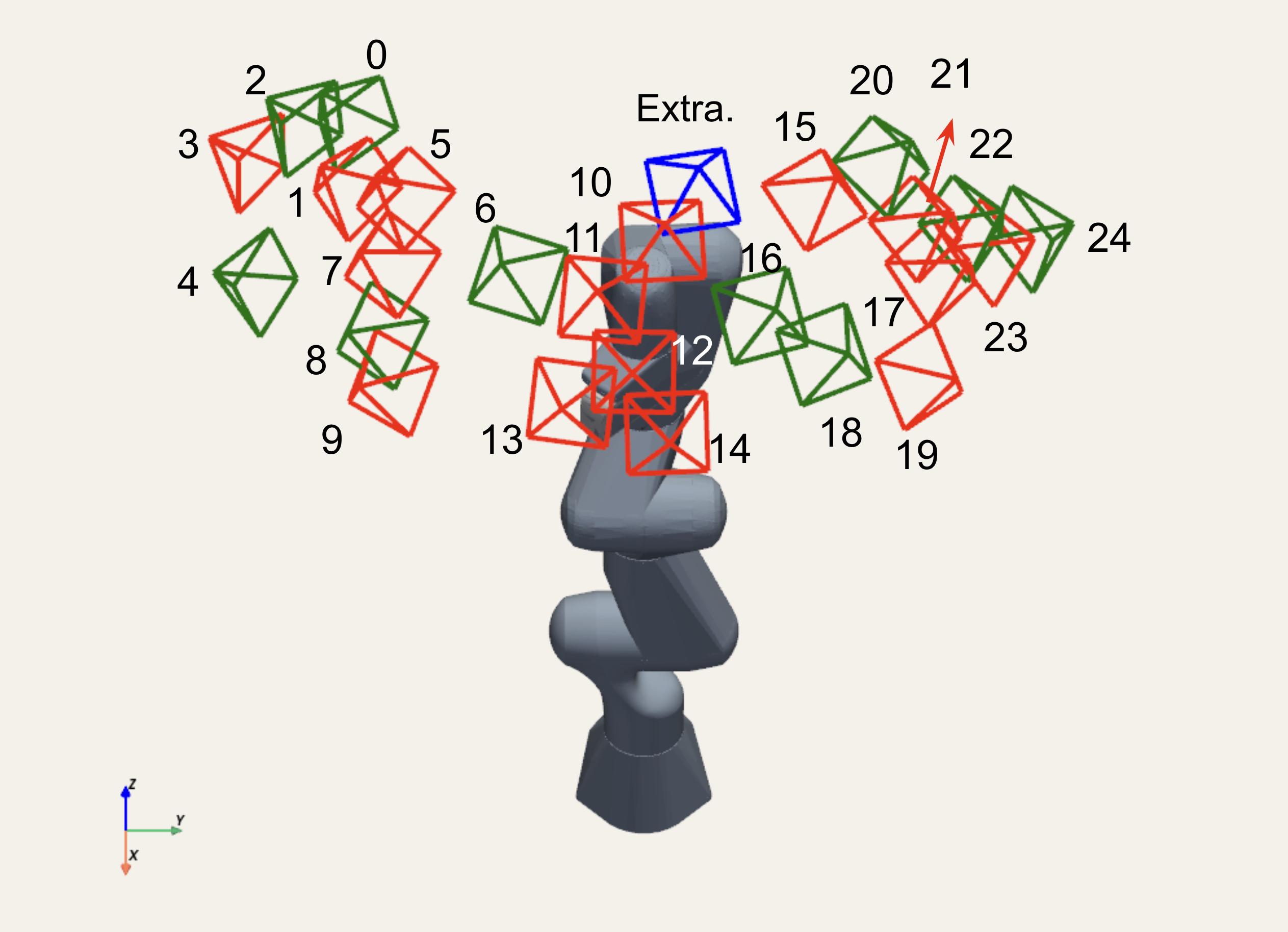}
    
  \end{subfigure}
  \caption{
  \textbf{Multi-View Camera Poses for Training and Evaluations.}
\textcolor{green!50!black}{Green} poses are used for training the encoder and policy (seen), and \textcolor{red!70!black}{red} poses are reserved for evaluation (unseen). An extrapolated viewpoint (beyond $5\times5$ grid) is generated starting from the \textcolor{blue!80!black}{blue} pose. 
  }
  \label{fig:25views}
\end{figure}

\begin{figure*}[htbp]
    \centering
    \begin{minipage}[c]{0.05\textwidth}
        \textbf{(a)}

    \end{minipage}%
    \begin{minipage}[c]{0.95\textwidth}
        \includegraphics[width=\linewidth]{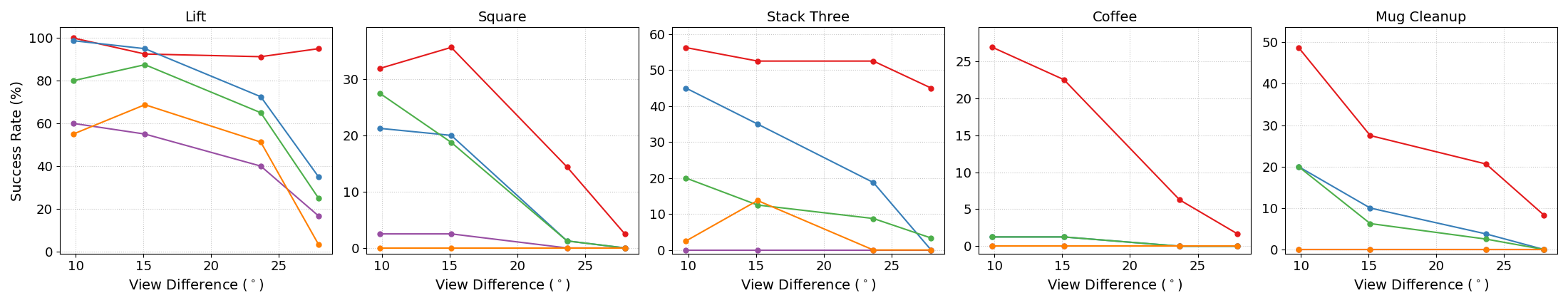} 
    \end{minipage}
    
    \vspace{1em}
    \begin{minipage}[c]{0.05\textwidth}
        \textbf{(b)}
     
    \end{minipage}%
    \begin{minipage}[c]{0.95\textwidth}
        \includegraphics[width=\linewidth]{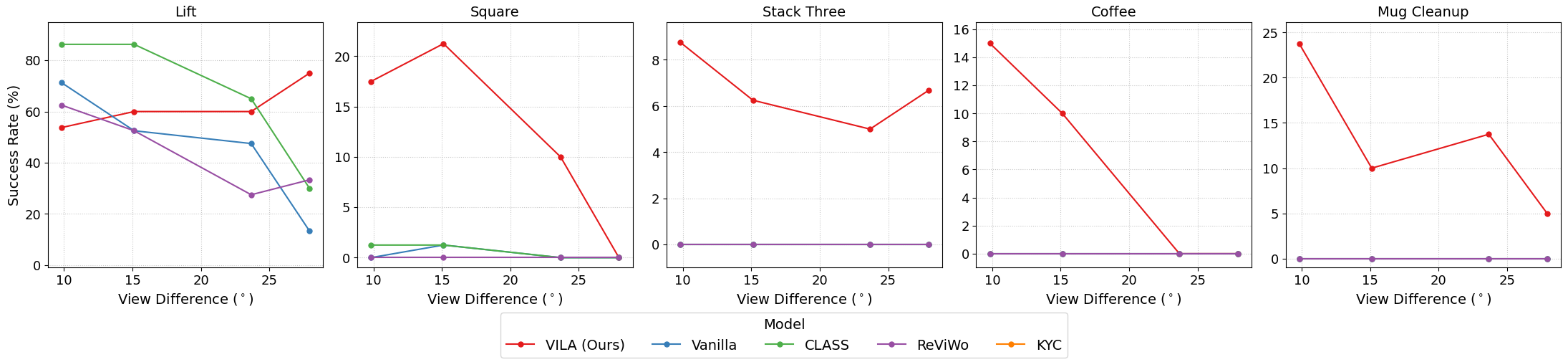}
    \end{minipage}
     \caption{
  \textbf{Unseen View Generalization vs. Viewpoint Difference.} Success rates (\%) (averaged over 20 episodes per each view) of VILA and baseline methods for view generalization under \textbf{(a)} \textit{fine-tuned} and \textbf{(b)} \textit{frozen} encoder settings. The success rates are shown with respect to angular differences from the training viewpoints. We evaluate 15 unseen camera viewpoints and, for each unseen view, compute its view difference to the closest training view among the 10 seen cameras, measured as the Euclidean norm in the (azimuth, elevation) space. Based on this, the 15 unseen views are sorted and partitioned into four groups (of sizes 4, 4, 4, and 3). On the x-axis, we plot the average nearest-view difference within each group, and on the y-axis we report the corresponding average success rate for that group.
}
    \label{fig:view_diff}
\end{figure*}

\subsection{Viewpoint Setups}
\label{sec:viewpoint_setups}

To train and evaluate view-robust policies, we construct a multi-view datasets by augmenting existing datasets with additional camera viewpoints. 
In simulation, we use five RoboSuite-based tasks~\cite{robosuite2020}: \subref{fig:lift}~\textbf{Lift}: simple block lifting and \subref{fig:square}~\textbf{Square}: \textbf{Pick \& Place} with precision tasks from RoboMimic~\cite{robomimic2021}, \subref{fig:stackthree}~\textbf{Stack Three}: multiple block stacking, \subref{fig:coffee}~\textbf{Coffee}: inserting pod in the coffee machine and closing, and \subref{fig:mugcleanup}~\textbf{Mug Cleanup}: opening the cabinet, placing mug in the cabinet, and closing tasks from MimicGen~\cite{mandlekar2023mimicgen}. 

For each trajectory, we treat the original \texttt{agentview} as a reference and define a $5\times5$ grid over azimuth offsets in $[-90^\circ, +90^\circ]$ and elevation offsets in $[-15^\circ, +15^\circ]$, partitioning each range into five uniform bins. 

We then sample one azimuth/elevation pair per grid cell as a relative offset from \texttt{agentview}, yielding 25 distinct camera poses and thus 25 view-augmented versions of every trajectory.
The resulting offsets ensure that both training and testing occur under non-trivial deviations from the original camera. 
From these 25 views, we fix 10 for training encoders and policies and reserve the remaining 15 for evaluation (Figure~\ref{fig:25views}). 
We also evaluate \emph{extrapolated} viewpoints by choosing a base pose whose azimuth and elevation lie outside this sampling range and then perturbing it by $\pm 2^\circ$ along both axes, giving 8 additional test views. 
Exact azimuth and elevation values for all views are provided in the Appendix.
Compared to prior multi-view setups used by our baselines~\citep{pang2025reviwo,lee2025class,jiang2025knowcameraisviewinvariant}, our benchmark covers a wide range of azimuth and elevation with fewer training views, leading to sparser camera coverage and a more challenging viewpoint generalization setting.

For real-world experiments, we use a self-collected single-view dataset of the SO-ARM101 robot~\cite{cadene2024lerobot} of two tasks: ~\subref{subfig:pnp} \textbf{Pick \& Place}: picking a block and placing inside the cup, and ~\subref{subfig:drawer} \textbf{Drawer}: putting a block in the drawer and closing it. 
Since capturing identical trajectories from many physical cameras is challenging, we follow prior work~\cite{chen2024roviaugrobotviewpointaugmentation,tian2025viewinvariantpolicylearningzeroshot} and apply ZeroNVS~\cite{zeronvs} to augment the original videos with additional viewpoints. 
We vary the camera azimuth in $\{-5^\circ, 0^\circ, +5^\circ\}$, the vertical translation in $\{-5\text{cm}, 0\text{cm}, +5\text{cm}\}$, and the optical-axis translation in $\{-5\text{cm}, 0\text{cm}, +5\text{cm}\}$, producing 27 views including the original. 
We select 4 views for training and evaluate on 3 held-out views in Figure~\ref{fig:real_world_views} that are disjoint from the training set.

\subsection{Unseen View Generalization}
\label{sec:unseen_view_generalization}

\textbf{Evaluation protocol.}
For each dataset and task, we first train an encoder and a visuomotor policy, and then evaluate generalization performance on seen and unseen viewpoints within the same task.
We use an image resolution of $64\times64$ and a latent dimension of $128$ for simulation tasks, whereas for real-world experiments, we use a resolution of $128\times128$ and a latent dimension of $512$.
All encoders are trained using the multi-view setup described in Sec.~\ref{sec:viewpoint_setups}.
For downstream control, we train a Diffusion Policy~\cite{chi2023diffusionpolicy} with identical hyperparameters across all baselines, and measure its success rate separately for each viewpoint.
Success rates are computed over multiple rollouts per view; additional details are reported in the Appendix.

We evaluate each representation in two downstream settings:
(i) a \emph{frozen} setting, where the encoder is fixed and only the policy parameters are trained, and
(ii) a \emph{fine-tuned} setting, where the encoder is updated jointly with the policy.
The frozen setting highlights the inherent viewpoint generalization of the learned representation itself, while the fine-tuned setting measures its effectiveness as a strong initialization for task-specific policy learning.

\paragraph{Baselines.}
All baselines share the same downstream policy, observation preprocessing, and training schedule; only the encoder pre-training strategy and the presence of camera-conditioning (Know Your Camera~\citep{jiang2025knowcameraisviewinvariant} only) differ.
This setup allows us to isolate the impact of the learned representation on unseen-view generalization.

\begin{itemize}    
    \item \textbf{Vanilla:} ImageNet-pretrained ResNet-18 encoder \citep{he2015deepresiduallearningimage, imagenet} used directly for policy learning. 
    
    \item \textbf{CLASS \citep{lee2025class}:} Scene-level encoder trained with a weighted InfoNCE loss based on GT action-sequence distances.
    
     \item \textbf{ReViWo \citep{pang2025reviwo}:} View-invariant scene representation learned via decomposition of multi-view observations.
    
    \item \textbf{Know Your Camera (KYC) \citep{jiang2025knowcameraisviewinvariant}:} Policy conditioned explicitly on camera parameters to aid cross-view generalization. 
\end{itemize}

\begin{table}[t!]
    \centering
    \caption{\textbf{View Generalization in Extrapolated Views in the Fine-tuned Setting.}
Success rates (\%) of fine-tuned policies on four simulation task evaluated under extrapolated camera poses (8 views).
For each task, we evaluate success over 20 episodes per view, reporting the average success rate across these views.
Note that the Coffee task is excluded because all methods failed under these extrapolated viewpoints.
}
    \label{tab:view_generalization_ft_extra}
    \resizebox{0.85\columnwidth}{!}{
    \begin{tabular}{l cccc}
        \toprule
        \multirow{2}{*}{\textbf{Model}} & \multirow{2}{*}{\textbf{Lift}} & \multirow{2}{*}{\textbf{Square}} & \textbf{Stack} & \textbf{Mug}\\
        &  &  & \textbf{Three} & \textbf{Cleanup}\\
        \midrule\midrule
        \rowcolor{blue!20} 
        VILA (Ours) & \textbf{93.10} & \textbf{3.10} & \textbf{35.00} & \textbf{11.85} \\
        Vanilla & 28.10 & 0.60 & 6.25 & 2.50 \\
        CLASS & 51.30 & 0.60 & 2.50 & 0.60 \\
        ReViWo & 8.80 & 0.00 & 0.00 & 0.00 \\
        KYC & 21.20 & 0.00 & 0.00 & 0.00 \\
        \bottomrule
    \end{tabular}
    }
\end{table}

\begin{table}[h]
    \centering
    \caption{\textbf{Real-World View Generalization.}
    Success rates (\%) on real-world tasks. We evaluate \textbf{Pick \& Place} on three unseen views and \textbf{Drawer} on two unseen views. Each setting consists of 10 episodes.
    }
    \label{tab:view_generalization_real_world}
    
    \resizebox{1.0\columnwidth}{!}{
    \begin{tabular}{l cccc c ccc} 
        \toprule
        & \multicolumn{4}{c}{\textbf{Pick \& Place}} && \multicolumn{3}{c}{\textbf{Drawer}} \\
        \cmidrule(lr){2-5} \cmidrule(lr){7-9}
        
        \textbf{Model} & \textbf{View 1} & \textbf{View 2} & \textbf{View 3} & \textbf{Avg.} && \textbf{View 1} & \textbf{View 2} & \textbf{Avg.} \\
        \midrule
        \midrule
        
        \rowcolor{blue!20}
        VILA (Ours) 
        & \textbf{70.00} & \textbf{80.00} & \textbf{40.00} & \textbf{63.33} && \textbf{80.00} & \textbf{90.00} & \textbf{85.00} \\
        
        Vanilla 
        & 0.00 & 0.00 & 10.00 & 3.33 && 0.00 & 0.00 & 0.00 \\
        
        CLASS 
        & 10.00 & 30.00 & 0.00 & 13.33 && 0.00 & 0.00 & 0.00 \\
        
        \bottomrule
    \end{tabular}
    }
\end{table}

\paragraph{Simulation results.}
Table~\ref{tab:view_generalization_combined} reports seen/unseen success rates and unseen/seen performance ratios (Rel.) across five simulated tasks in both \emph{frozen} and \emph{fine-tuned} settings. With fine-tuning, VILA attains the best unseen-view success on all tasks, including the harder tasks, while in the frozen setting it is the only method that maintains non-trivial performance on tasks where other approaches often collapse to near-zero under unseen views. Figure~\ref{fig:view_diff} further breaks down unseen-view performance as a function of viewpoint difference from seen views: across both settings, VILA consistently outperforms all baselines and its success degrades much more slowly as the gap from the nearest training camera increases. 
Table~\ref{tab:view_generalization_ft_extra} shows that these gaps widen under extrapolated camera poses, with VILA retaining meaningful success rates while baselines largely fail.

\paragraph{Real-world results.}
Table~\ref{tab:view_generalization_real_world} presents the performance on real-world \textbf{Pick \& Place} and \textbf{Drawer} tasks under novel viewpoints. 
VILA achieves average success rates of $63.3\%$ and $85.0\%$, respectively. 
In comparison, baseline methods show significantly lower performance, averaging below $13.3\%$ on \textbf{Pick \& Place} and recording $0\%$ on the \textbf{Drawer} task. 
These results indicate that VILA offers improved robustness to viewpoint changes compared to the baselines.

\begin{figure}
  \centering
  \begin{subfigure}{\columnwidth}
    \centering 
    \includegraphics[width=\columnwidth]{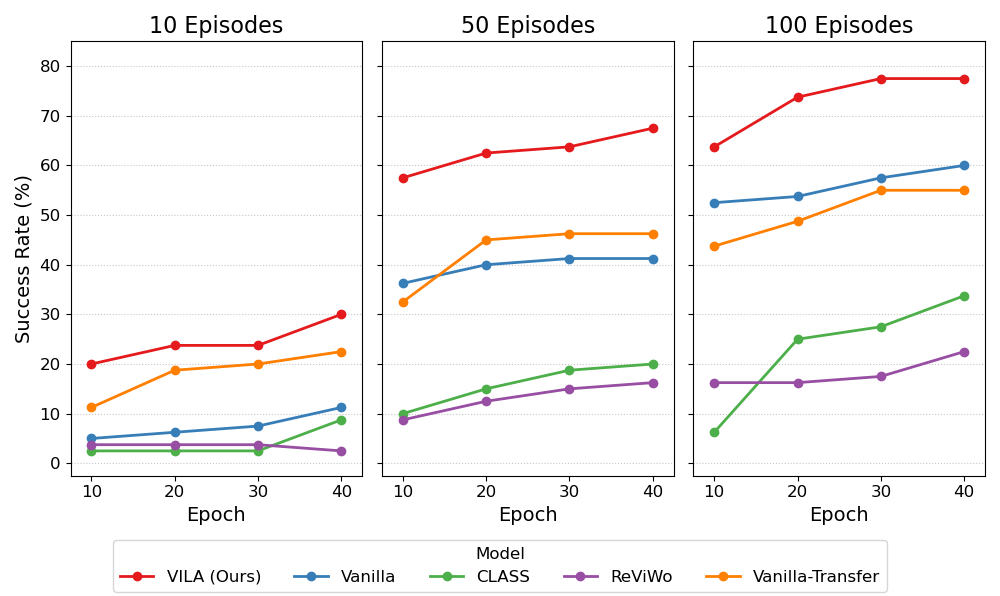}
    \label{fig:my_sub_figure}
  \end{subfigure}
  \caption{
    \textbf{Unseen Task Adaptation.} Success rates (\%) of VILA and baseline methods when adapting from Stack Three to Coffee. For each data point, we report the average success over 40 episodes using the training view only, plotted as a function of the number of labeled trajectories and training epochs.
  }
  \label{fig:task_adaptation_combined} 
\end{figure}

\subsection{Unseen Task Adaptation}
\label{sec:unseen_task}

\paragraph{Evaluation protocol.}
To investigate whether representations learned on one dataset provide useful priors for other tasks, we transfer encoders trained on the Stack Three task to a new Coffee task under a single-view setup (i.e., policy is trained and evaluated on the single identical view).
Concretely, we take each encoder pre-trained on Stack Three in the multi-view setting (Sec.~\ref{sec:viewpoint_setups}) and use it to initialize a visuomotor policy on Coffee, then fine-tune the encoder and policy jointly using only a small subset of the Coffee demonstrations. Note that policies for Coffee task are trained with 1{,}000 episodes in Section~\ref{sec:unseen_view_generalization}.
For each encoder, we train single-view policies from two different camera viewpoints and report performance as the average success rate across these views.
This setup allows us to assess both cross-task transfer and whether each encoder provides a viewpoint-generalized prior that carries over to a new task.

\paragraph{Baselines.}
The Vanilla baseline denotes training a policy directly on the Coffee task in the single-view setting, using only the chosen Coffee subset and an ImageNet-pretrained ResNet encoder, without any multi-view or cross-task pre-training.
The Vanilla-Transfer baseline instead first fine-tunes the Vanilla encoder on the Stack Three multi-view task (as in our unseen-view generalization experiments), and then uses this fine-tuned encoder as initialization for subsequent single-view training on Coffee.
All other methods follow the same protocol: we start from their respective encoders pre-trained on the Stack Three multi-view dataset and then fine-tune them on the single-view Coffee task.

\paragraph{Results.}
Figure~\ref{fig:task_adaptation_combined} summarizes unseen task adaptation on Coffee under different labeled data budgets. 
Across all budgets, VILA provides a stronger prior than the Vanilla baseline, while other encoders often match or even underperform Vanilla trained from scratch. 
Thus, not all multi-view pre-training is helpful for cross-task adaptation: enforcing scene-level invariance can produce less transferable priors that might overfit to task-specific appearances, whereas VILA’s latent-action representation yields a viewpoint-generalized and dynamics-centered prior that remains useful even with limited Coffee data.

\section{Discussion}
\label{sec:discussion}
\subsection{Ablation Studies}
\label{sec:ablation}

We conduct a comprehensive ablation study to validate the key design choices of VILA.
Unless otherwise noted, all ablations are performed on the Lift task in the fine-tuned setting, and we report success rates on unseen views.
Our default configuration uses an action-guided Weighted InfoNCE loss combined with a structural distance-alignment loss, L2 distance between action sequences for weight and cosine similarity for structural alignment, a latent action dimension of 128, and a random offset sampling range of 10 steps (Sec.~\ref{sec:action_contrastive}).
The results are summarized in Table~\ref{tab:ablation}.

\begin{table}[t!]
    \centering
    \caption{\textbf{Ablation study of VILA components.} Seen and Unseen denote success rates (\%) averaged over 20 episodes across the 10 training views and 15 held-out views on the Lift task in the fine-tuned setting.}
    \label{tab:ablation}
    
    \resizebox{\columnwidth}{!}{
    \begin{tabular}{l rrr}
        \toprule
        \textbf{Method} & \textbf{Seen} & \textbf{Unseen} & \textbf{Rel.} \\
        \midrule\midrule
        \rowcolor{blue!20}
        \textbf{VILA (Ours)} & \textbf{99.50} & \textbf{94.70} & \textbf{95.18} \\
        \midrule

        \multicolumn{4}{l}{\textit{Loss function (default: }$\mathcal{L}_{\text{LA}} \;\&\; \mathcal{L}_{\text{W-NCE}} \;\&\; \mathcal{L}_{\text{struct}}$)} \\

        \quad $\mathcal{L}_{\text{LA}} \;\&\; \mathcal{L}_{\text{W-NCE}} \;\&\; \mathcal{L}_{\text{CKA}}$ & 98.50 & 92.00 & 93.40 \\
        \quad $\mathcal{L}_{\text{LA}} \;\&\; \mathcal{L}_{\text{W-NCE}}$ & 99.00 & 91.70 & 92.63 \\
        \quad $\mathcal{L}_{\text{LA}} \;\&\; \mathcal{L}_{\text{NCE}}$ & 95.50 & 90.00 & 94.24 \\
        \quad $\mathcal{L}_{\text{LA}} \;\&\; \mathcal{L}_{\text{struct}}$ & 99.50 & 84.30 & 84.72 \\
        \quad $\mathcal{L}_{\text{LA}}$ & 93.00 & 79.50 & 85.48 \\
        \quad $\mathcal{L}_{\text{LA}} \;\&\; \mathcal{L}_{\text{W-NCE}} \;\&\; \mathcal{L}_{\text{struct}} \;\&\; \mathcal{L}_{\text{act}}$ & 98.00 & 86.70 & 88.47 \\
        \midrule

        \multicolumn{4}{l}{\textit{Action-sequence distance}} \\
        \quad Dynamic Time Warping (DTW) & 96.40 & 89.20 & 92.53 \\
        \midrule

        \multicolumn{4}{l}{\textit{Offset sampling strategy (default: random offset 1--10)}} \\
        \quad Random offset 1--16 & 99.50 & 93.70 & 94.17 \\
        \quad Random offset 1--5 & 98.50 & 92.30 & 93.71 \\
        \quad Fixed offset 10 & 97.50 & 90.00 & 92.31 \\
        \midrule

        \multicolumn{4}{l}{\textit{Latent action dimension (default: 128)}} \\
        \quad 32 dimensions & 98.00 & 88.00 & 89.80 \\
        \quad 64 dimensions & 97.00 & 84.70 & 87.32 \\
        \quad 256 dimensions & 98.00 & 88.70 & 90.51 \\
        \quad 512 dimensions & 96.00 & 90.00 & 93.75 \\
        \midrule

        \multicolumn{4}{p{\columnwidth}}{ 
        \footnotesize
        \textbf{Loss term definitions.}
        $\mathcal{L}_{\text{NCE}}$: standard InfoNCE with positives from the same transition and negatives from others;\newline
        $\mathcal{L}_{\text{CKA}}$: RBF-kernel CKA-based structural alignment of latent actions and GT-actions;\newline
        $\mathcal{L}_{\text{act}}$: auxiliary action-prediction loss.
    } \\
        \bottomrule
    \end{tabular}
    }
\end{table}

\paragraph{Loss function.}
Removing the action-based weighting with standard InfoNCE or omitting the structural loss both degrade unseen-view success compared to the full objective, indicating that the two components are complementary. 
Using the base latent-action loss alone performs even worse, confirming that neither the structural loss nor the action-based weighting is helpful. 
Alternative global regularizers based on distance-matrix or CKA-style alignment also underperform our distance-based structural loss. 
Adding an auxiliary action-regression head further harms unseen-view performance, suggesting that actions are more effective as soft similarity supervision than as direct regression targets.

\paragraph{Offset sampling strategy.}
Sampling temporal offsets uniformly from $\{1,\dots,10\}$ when constructing action sequences yields the best unseen-view generalization. 
Smaller or larger maximum offsets, or a fixed offset, consistently degrade performance, suggesting that a moderate offset range captures dynamics most effectively.

\paragraph{Distance metric.}
Replacing our L2 distance in weighted constrastive learning and cosine similiarity in structural alignment with Dynamic Time Warping (DTW)~\cite{dtw} lowers unseen-view success. 
Unlike prior work that applies DTW to longer trajectories (e.g., 16 steps in CLASS~\cite{lee2025class}), our sequences of up to 10 steps already work well with L2. 
Since we resample sequence lengths (1–10) at every batch, precomputing exact DTW distances is impractical, so we instead use Soft-DTW~\cite{cuturi2018softdtwdifferentiablelossfunction}, whose approximation error and higher computational cost do not translate into better performance in our setting.

\paragraph{Latent action dimension.}
A latent dimension $128$ gives the best unseen-view generalization. 
Both smaller and much larger latent spaces hurt performance, suggesting that $128$ offers a good trade-off between expressivity and regularization.

\subsection{Representation Quality Analysis}

We evaluate the quality of the representation using entropy-based metrics that capture the viewpoint invariance and dynamics-aware semantics. 
For each encoder, we sample $12{,}500$ transitions (500 per view) from the multi-view dataset, extract features before and after policy fine-tuning, and for each feature compute its $k=50$ nearest neighbors (L2) and the corresponding entropies.

\begin{table}[h]
    \centering
    \caption{\textbf{View and Action Entropy.}
Entropy-based analysis of representation quality before and after policy fine-tuning in Lift dataset.
“Seen” and “Unseen” denote view entropies (higher is better, ↑) computed over the 25 views, respectively, while “Action” denotes action entropy (lower is better, ↓) based on 10 clustered action classes.
The upper-bound row corresponds to the entropy of a uniform distribution over 25 views (Seen/Unseen) and 10 action clusters (Action).
}
    \label{tab:view_action_entropy}
    \resizebox{0.85\columnwidth}{!}{
    \begin{tabular}{l rrrr}
        \toprule
        \textbf{Model} & \textbf{Seen (↑)} & \textbf{Unseen (↑)} & \textbf{Action (↓)} \\
        \midrule\midrule
        Upper Bound & 3.219 & 3.219 & 2.303 \\
        \midrule\midrule
        \multicolumn{4}{l}{\textbf{Before Fine-tuning}} \\
        \midrule
        \rowcolor{blue!20}
        VILA (Ours) & \textbf{2.757} & \textbf{2.495} & \textbf{0.934} \\
        CLASS & 2.641 & 2.159 & 1.041 \\
        ReViWo & 2.479 & 1.83 & 1.132 \\
        \midrule\midrule
        \multicolumn{4}{l}{\textbf{Fine-Tuned}} \\
        \midrule
        \rowcolor{blue!20}
        VILA (Ours) & \textbf{2.803} & \textbf{2.548} & \textbf{0.829} \\
        Vanilla & 2.707 & 2.121 & 1.445 \\
        CLASS & 2.624 & 1.933 & 1.006 \\
        ReViWo & 2.597 & 2.244 & 1.177 \\
        KYC & 2.149 & 1.793 & 1.254 \\
        \bottomrule
    \end{tabular}
    }
\end{table}

\paragraph{View entropy.}
View entropy measures how mixed camera views are in each feature's local neighborhood. 
For each feature, we take the distribution of view IDs among its $k$ nearest neighbors, compute the Shannon entropy~\citep{shannon} of this categorical distribution, and average over all features. 
\emph{Higher} view entropy means neighbors are drawn more uniformly from the 25 views, indicating a more view-invariant representation.

\paragraph{Action entropy.}
To check that invariance does not harm dynamics-aware semantics, we define an action entropy that measures how consistently features group similar future dynamics. 
We cluster all 10-step GT action sequences $\{a_t,\dots,a_{t+9}\}$ into $K=10$ action classes using $k$-means and assign each observation $o_t$ the label $c_t$ of its associated sequence. 
For each feature, we look at the class labels of its $k$ nearest neighbors, compute the entropy over this distribution, and average across features. 
Here, \emph{lower} action entropy is better, indicating that visual features tend to cluster according to similar action outcomes.

\paragraph{Results.}
Table~\ref{tab:view_action_entropy} reports view and action entropy on the Lift dataset (other tasks are in the Appendix).
VILA achieves the \emph{highest view entropy} on both seen and unseen views, indicating a strongly view-invariant representation, while also obtaining the \emph{lowest action entropy}. 
This shows that enforcing invariance on \emph{dynamics} rather than entire scenes lets VILA simultaneously achieve strong viewpoint mixing and tight clusters of features with similar action outcomes.

\begin{figure}
  \centering
  
  \begin{subfigure}{1\columnwidth}
    \includegraphics[width=\linewidth]{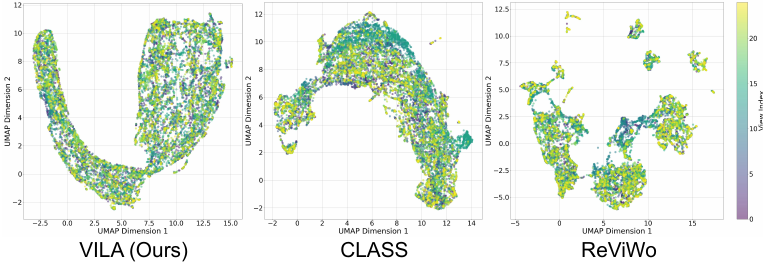}
    \caption{\textbf{Before Policy Training}}
    \label{fig:umap_views_before}
  \end{subfigure}
  
  \begin{subfigure}{1\columnwidth}
    \includegraphics[width=\linewidth]{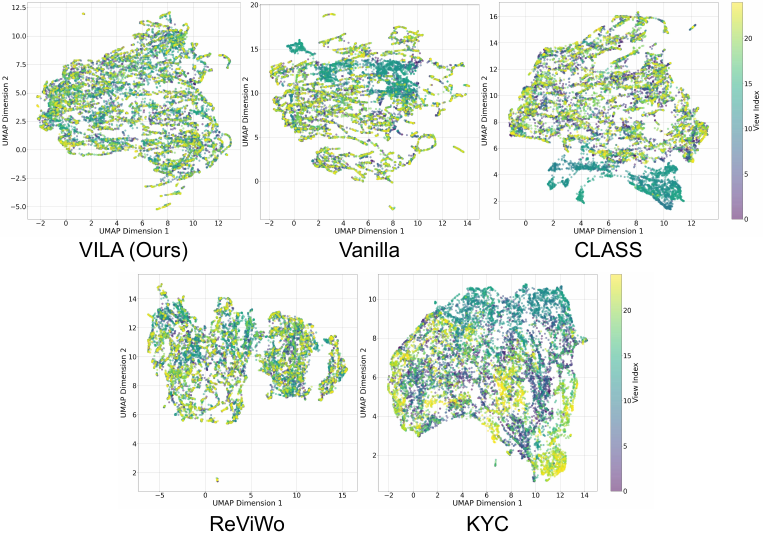}
    \caption{\textbf{After Policy Training}}
    \label{fig:umap_views_after}
  \end{subfigure}

  \caption{
    \textbf{UMAP of encoder representations across 25 views.}
    On Lift, baselines show distinct clusters for unseen views (especially for views 10–14), 
    whereas VILA representations are more uniformly mixed across views, indicating stronger 
    view invariance both before and after policy training.
  }
  \label{fig:umap}
\end{figure}

\paragraph{UMAP plots.}
We visualize the learned representations with UMAP~\citep{mcinnes2020umapuniformmanifoldapproximation}. 
Figures~\ref{fig:umap_views_before} and~\ref{fig:umap_views_after} show 2D embeddings colored by view index. For baselines, unseen views (especially Views 10--14) form separate clusters, whereas VILA intermingles them with seen views. 
This pattern matches our entropy analysis that VILA has the highest view entropy.
Additional UMAP plots for latent actions and other datasets are in the Appendix.

\section{Conclusion}

We tackle viewpoint robustness in visuomotor control with VILA, a pre-training framework that enforces invariance on latent action instead of scene-level visual features. 
Building on latent action models with an action-guided contrastive loss and structure alignment, VILA learns a latent space that is both view-invariant and aligned with control dynamics, leading to consistent gains in unseen-view generalization and data-efficient unseen task adaptation across five simulated tasks and a real-world SO-ARM setup. These suggest that targeting invariance at the level of dynamics is a promising direction for robust visuomotor policies under camera changes.

\paragraph{Limitations.}
Our experiments assume access to multi-view observations of a given task, which are straightforward to generate in simulation but more involved to collect in real-world setups. 
In the real-robot setting, we follow prior work and use ZeroNVS to obtain additional viewpoints, so part of the observed robustness may reflect the behavior of the underlying NVS model. 
Finally, we primarily study robustness to camera pose; extending the same action-guided invariance principle to other sources of variation (e.g., lighting, backgrounds, object appearance) is a natural next step toward more broadly robust visuomotor policies.

{
    \small
    \bibliographystyle{ieeenat_fullname}
    \bibliography{main}
}

\clearpage
\setcounter{page}{1}
\maketitlesupplementary
\appendix

\section{Implementation Details}
\label{sec:appendix_conf}
We provide the implementation details used for view generation, NVS generation, VILA training, baselines and diffusion policy training. 

\subsection{View Configurations}
For reproducibility, Tables~\ref{tab:camera-poses-main} and
\ref{tab:camera-poses-extra} list the exact world-frame camera poses
(positions and MuJoCo quaternions) for all views used in this paper.

\subsection{ZeroNVS Configurations}
 We utilized the official open-source implementation of ZeroNVS\footnote{\url{https://github.com/kylesargent/ZeroNVS}} and the pretrained checkpoint to generate novel views in our real-world experiments using SO-ARM101. Configurations for this process is as Table \ref{tab:nvs_params}. We created 26 more combinations of novel views of our dataset and selected 4 views of azimuth, vertical and optical axis translation of $(0.0^\circ, -5\text{cm}, 5\text{cm}), (5^\circ, 0\text{cm}, -5\text{cm})$, $(-5^\circ, 5\text{cm}, 0\text{cm})$ and the original data representing $(0.0^\circ, 0.0\text{cm}, 0.0\text{cm})$ to cover all variations of each axis.

\subsection{VILA Configurations}
We train VILA with a two-stage pipeline: (i) latent action pre-training from multi-view videos, and (ii) latent behavior cloning in the learned latent space. Unless otherwise specified, we use a single configuration across all datasets and tasks, summarized in Table~\ref{tab:vila_hparams}. 
\begin{table}
\centering
\scriptsize
\setlength{\tabcolsep}{4pt}
\begin{tabular}{lrrrrrrrr}
\toprule
View & $x$ & $y$ & $z$ & $q_w$ & $q_x$ & $q_y$ & $q_z$ \\
\midrule
View 00 & 0.0981 & -0.4091 & 1.4249 & 0.9451 & 0.3051 & 0.0361 & 0.1117 \\
View 01 & 0.2021 & -0.4080 & 1.3951 & 0.9159 & 0.3304 & 0.0773 & 0.2144 \\
View 02 & 0.0473 & -0.4811 & 1.3677 & 0.9293 & 0.3662 & 0.0180 & 0.0456 \\
View 03 & 0.0204 & -0.5587 & 1.2741 & 0.8937 & 0.4483 & 0.0082 & 0.0163 \\
View 04 & 0.2175 & -0.5482 & 1.2234 & 0.8577 & 0.4786 & 0.0915 & 0.1639 \\
View 05 & 0.2523 & -0.3380 & 1.4240 & 0.9029 & 0.2925 & 0.0971 & 0.2998 \\
View 06 & 0.3973 & -0.1771 & 1.4131 & 0.7947 & 0.2686 & 0.1743 & 0.5157 \\
View 07 & 0.3278 & -0.3572 & 1.3663 & 0.8665 & 0.3429 & 0.1335 & 0.3374 \\
View 08 & 0.3985 & -0.3722 & 1.2939 & 0.8271 & 0.3965 & 0.1722 & 0.3592 \\
View 09 & 0.4504 & -0.3576 & 1.2489 & 0.7957 & 0.4215 & 0.2036 & 0.3843 \\
View 10 & 0.3738 & 0.0263 & 1.4587 & 0.6572 & 0.1817 & 0.1950 & 0.7051 \\
View 11 & 0.4394 & -0.0577 & 1.4061 & 0.7102 & 0.2465 & 0.2162 & 0.6230 \\
View 12 & 0.5130 & -0.0142 & 1.3349 & 0.6579 & 0.2846 & 0.2769 & 0.6400 \\
View 13 & 0.5352 & -0.1022 & 1.2945 & 0.6950 & 0.3327 & 0.2752 & 0.5749 \\
View 14 & 0.5651 & 0.0328 & 1.2635 & 0.6105 & 0.3135 & 0.3322 & 0.6470 \\
View 15 & 0.2890 & 0.2414 & 1.4574 & 0.4081 & 0.1136 & 0.2428 & 0.8727 \\
View 16 & 0.4437 & 0.1645 & 1.3781 & 0.5336 & 0.2036 & 0.2926 & 0.7669 \\
View 17 & 0.3112 & 0.4094 & 1.3336 & 0.2929 & 0.1271 & 0.3773 & 0.8693 \\
View 18 & 0.4612 & 0.2553 & 1.3178 & 0.4627 & 0.2092 & 0.3549 & 0.7850 \\
View 19 & 0.4271 & 0.3932 & 1.2399 & 0.3535 & 0.1908 & 0.4350 & 0.8059 \\
View 20 & 0.2071 & 0.3366 & 1.4443 & 0.2611 & 0.0775 & 0.2737 & 0.9225 \\
View 21 & 0.2687 & 0.3871 & 1.3800 & 0.2793 & 0.1059 & 0.3384 & 0.8923 \\
View 22 & 0.2279 & 0.4535 & 1.3414 & 0.2124 & 0.0903 & 0.3806 & 0.8955 \\
View 23 & 0.2285 & 0.4949 & 1.2941 & 0.1935 & 0.0927 & 0.4220 & 0.8808 \\
View 24 & 0.1605 & 0.5482 & 1.2553 & 0.1258 & 0.0657 & 0.4585 & 0.8773 \\
\bottomrule
\end{tabular}
\caption{\textbf{Exact Camera Poses for the 25 Main Views. }
Positions are given in the world frame (MuJoCo’s default metric units, i.e., meters),
and orientations are MuJoCo quaternions $(q_w, q_x, q_y, q_z)$.
}
\label{tab:camera-poses-main}
\end{table}

\begin{table}
\centering
\scriptsize
\setlength{\tabcolsep}{4pt}
\begin{tabular}{lrrrrrrrr}
\toprule
View & $x$ & $y$ & $z$ & $q_w$ & $q_x$ & $q_y$ & $q_z$ \\
\midrule
Extra View 00 & 0.2902 & 0.0585 & 1.5040 & 0.6269 & 0.1256 & 0.1568 & 0.7527 \\
Extra View 01 & 0.3111 & 0.0577 & 1.4936 & 0.6240 & 0.1391 & 0.1672 & 0.7505 \\
Extra View 02 & 0.3316 & 0.0570 & 1.4825 & 0.6209 & 0.1525 & 0.1776 & 0.7481 \\
Extra View 03 & 0.2880 & 0.0686 & 1.5040 & 0.6137 & 0.1228 & 0.1590 & 0.7635 \\
Extra View 04 & 0.3294 & 0.0686 & 1.4825 & 0.6078 & 0.1494 & 0.1803 & 0.7588 \\
Extra View 05 & 0.2854 & 0.0786 & 1.5040 & 0.6003 & 0.1200 & 0.1611 & 0.7741 \\
Extra View 06 & 0.3063 & 0.0793 & 1.4936 & 0.5975 & 0.1331 & 0.1720 & 0.7718 \\
Extra View 07 & 0.3268 & 0.0800 & 1.4825 & 0.5945 & 0.1462 & 0.1829 & 0.7693 \\
\bottomrule
\end{tabular}
\caption{\textbf{Exact Camera Poses for the Extrapolated Views. }
Positions are given in the world frame (MuJoCo’s default metric units, i.e., meters),
and orientations are MuJoCo quaternions $(q_w, q_x, q_y, q_z)$.}
\label{tab:camera-poses-extra}
\end{table}

\begin{table}[h]
    \centering
    \caption{\textbf{ZeroNVS Hyperparameters.} Configuration for ZeroNVS for real-world data augmentation.}
    \label{tab:nvs_params}
    \resizebox{0.7\linewidth}{!}{
    \begin{tabular}{lc}
        \toprule
        \textbf{Hyperparameter} & \textbf{Value} \\
        \midrule
        Noise Scheduler & DDIM \\
        Inference Steps & 50 \\
        Guidance Scale & 7.5 \\
        FOV deg & $13.0^\circ$ \\
        Default Elevation & $-10.0^\circ$ \\
        Default Azimuth & $0.0^\circ$ \\
        Default Distance & $1.0\text{m}$ \\
        Translation Azimuth & $\{-5.0, 0.0, 5.0\}^\circ$ \\
        Translation Vertical & \{-5.0, 0.0, 5.0\}\text{cm} \\
        Translation Optical & \{-5.0, 0.0, 5.0\}\text{cm} \\
        \bottomrule
    \end{tabular}
    }
\end{table}

\begin{table}[h]
    \centering
    \caption{\textbf{VILA Training Hyperparameters.} We use a two-stage training pipeline: Stage 1 latent action pre-training and Stage 2 latent behavior cloning.}
    \label{tab:vila_hparams}
    \resizebox{0.7\linewidth}{!}{
    \begin{tabular}{lc}
        \toprule
        \textbf{Hyperparameter} & \textbf{Value} \\
        \midrule
        \multicolumn{2}{l}{\textbf{Stage 1: Latent Action Pre-training}} \\
        Optimizer & AdamW \\
        Learning Rate & $1 \times 10^{-4}$ \\
        Latent Dimension ($D_z$) & 128 \\
        Prediction Horizon ($K$) & $1 \sim 10$ \\
        Time Indices per Batch ($N$) & 16 \\
        Views per Batch ($V$) & 8 \\
        Image Resolution & $64 \times 64$ \\
        Distance Type & $L_2$ \\
        InfoNCE Temperature ($\tau$) & 1.0 \\
        $\lambda_1$ ($\mathcal{L}_{\text{W-NCE}}$) & 1.0 \\
        $\lambda_2$ ($\mathcal{L}_{\text{struct}}$) & 1.0 \\
        Weighting Temperature ($\beta$) & 0.001 \\
        Epochs & 100 \\
        Gradient Clipping Norm & 1.0 \\
        Target EMA Coef. & 0.001 \\
        Target Update Interval (Epochs) & 1 \\
        IDM Head Hidden Dim. & 1024 \\
        FDM Head Hidden Dim. & 1024 \\
        \midrule
        \multicolumn{2}{l}{\textbf{Stage 2: Latent Behavior Cloning}} \\
        Optimizer & AdamW \\
        Learning Rate & $5 \times 10^{-5}$ \\
        BC Batch Size & 256 \\
        Prediction Horizon ($K$) & 10 \\
        Image Resolution & $64 \times 64$ \\
        Epochs & 100 \\
        \bottomrule
    \end{tabular}
    }
\end{table}

\subsection{Baseline Implementations}
\paragraph{CLASS.}
We adapt CLASS from the official open-source implementation\footnote{\url{https://github.com/sean1295/CLASS}} to our setting by applying its contrastive objective on top of our Stage-1 encoder.
To ensure a fair comparison, we match the configuration used in VILA: an image resolution of $64 \times 64$ and latent dimension of 128 for simulation, and $128 \times 128$ with a dimension of 512 for real-world experiments.

\paragraph{ReViWo.}
For ReViWo, we follow the official implementation\footnote{\url{https://github.com/Trevor-emt/Reviwo}}. 
We use input image sizes of $64 \times 64$ for simulation and $128 \times 128$ for real-world tasks. 
Since ReViWo is ViT-based, we adjust the token and hidden dimensions so that the final representation dimension matches 128 for simulation and 512 for the real-world setting, ensuring that all methods use the same latent dimensionality as VILA.

\paragraph{Know Your Camera.}
We adapt Know Your Camera (KYC) from official open-source implementation\footnote{\url{https://github.com/ripl/CamPoseOpensource}} to our setting. 
This method explicitly conditions the policy on camera extrinsic parameters by generating per-pixel 6-dimensional Plücker ray from the camera parameters. We concatenate the Plücker rays into the channel of the original image as suggested in the paper to apply the method to the diffusion policy.

\subsection{Diffusion Policy Configurations}
For simulation experiments, we used the official open-source implementation\footnote{\url{https://github.com/ARISE-Initiative/robomimic}} of diffusion policy in Robomimic~\cite{robomimic2021}. 
The main hyperparameters are summarized in Table~\ref{tab:policy_params_sim}. 
All methods (ours and baselines) use the same diffusion-policy architecture and training settings.

For real-world experiments, we utilized the official open-source implementation of the diffusion policy in LeRobot\footnote{\url{https://github.com/huggingface/lerobot}}. Configurations for this process are in Table \ref{tab:policy_params}.

\begin{table}[h]
    \centering
    \caption{\textbf{Diffusion Policy Hyperparameters for Simulations.} Configuration for the diffusion policy used for policy training in simulations.}
    \label{tab:policy_params_sim}
    \resizebox{0.9\linewidth}{!}{
    \begin{tabular}{lc}
        \toprule
        \textbf{Hyperparameter} & \textbf{Value} \\
        \midrule
        Learning Rate & $1 \times 10^{-4}$ \\
        Encoder Learning Rate (In Fine-tune Setting) & $1 \times 10^{-5}$ \\
        Optimizer & AdamW \\
        Optimizer Scheduler & Cosine \\
        Training Epochs (In Fine-tune setting) & 50 \\
        Training Epochs (In Frozen setting) & 100 \\
        Steps per Epoch & 1{,}000 \\
        Warmup Steps & 500 \\
        Batch Size (for Lift) & 100 \\
        Batch Size (for others) & 1{,}000 \\
        Image Resolution & $64 \times 64$ \\
        Observation Horizon & 1 \\
        Prediction Horizon & 16 \\
        Action Horizon & 8 \\
        Noise Scheduler & DDIM \\
        Beta Scheduler & squaredcos\_cap\_v2 \\
        Diffusion Training Steps & 100 \\
        Diffusion Inference Steps & 10 \\
        \bottomrule
    \end{tabular}
    }
\end{table}

\begin{table}[h]
    \centering
    \caption{\textbf{Diffusion Policy Hyperparameters for Real-World.} Configuration for the Diffusion Policy for policy finetuning in real-world experiments.}
    \label{tab:policy_params}
    \resizebox{0.7\linewidth}{!}{
    \begin{tabular}{lc}
        \toprule
        \textbf{Hyperparameter} & \textbf{Value} \\
        \midrule
        Learning Rate & $1 \times 10^{-4}$ \\
        Encoder Learning Rate & $1 \times 10^{-5}$ \\
        Optimizer & AdamW \\
        Optimizer Scheduler & Cosine \\
        Training Steps & 200,000 \\
        Warmup Steps & 500 \\
        Batch Size & 64 \\
        Image Resolution & $128\times128$ \\
        Observation Horizon & 2 \\
        Prediction Horizon & 16 \\
        Action Horizon & 16 \\
        Noise Scheduler & DDIM \\
        Beta scheduler & squaredcos\_cap\_v2 \\
        Diffusion Training Steps & 100 \\
        Diffusion Inference Steps & 10 \\
        \bottomrule
    \end{tabular}
    }
\end{table}

\section{Additional Experimental Results}
\subsection{Entropy Results}
Tables~\ref{tab:view_action_entropy_square}--\ref{tab:view_action_entropy_mug} report the same view and action entropy metrics on the remaining tasks (Square, Stack Three, Coffee, and Mug Cleanup).
Across all four datasets, we observe the same qualitative trend as in Lift: VILA consistently attains the highest view entropy on both seen and unseen views, while achieving the lowest action entropy among all methods, both before and after policy fine-tuning.

\begin{table}[h]
    \centering
    \caption{\textbf{View and Action Entropy (Square).}
    Entropy-based analysis of representation quality before and after policy fine-tuning in the Square dataset.
    “Seen” and “Unseen” denote view entropies (higher is better, ↑) computed over the 25 views, while “Action” denotes action entropy (lower is better, ↓) based on 10 clustered action classes.
    The upper-bound row corresponds to the entropy of a uniform distribution over 25 views (Seen/Unseen) and 10 action clusters (Action).
    }
    \label{tab:view_action_entropy_square}
    \resizebox{0.85\columnwidth}{!}{
    \begin{tabular}{lrrr}
        \toprule
        \textbf{Model} & \textbf{Seen (↑)} & \textbf{Unseen (↑)} & \textbf{Action (↓)} \\
        \midrule\midrule
        Upper Bound & 3.219 & 3.219 & 2.303 \\
        \midrule\midrule
        \multicolumn{4}{l}{\textbf{Before Fine-tuning}} \\
        \midrule
        \rowcolor{blue!20}
        VILA (Ours) & \textbf{2.894} & \textbf{2.705} & \textbf{0.631} \\
        CLASS & 2.607 & 1.982 & 0.861 \\
        ReViWo & 2.325 & 1.748 & 1.316 \\
        \midrule\midrule
        \multicolumn{4}{l}{\textbf{Fine-Tuned}} \\
        \midrule
        \rowcolor{blue!20}
        VILA (Ours) & \textbf{2.917} & \textbf{2.712} & \textbf{0.598} \\
        Vanilla & 2.607 & 1.982 & 0.802 \\
        CLASS & 2.449 & 1.57 & 0.912 \\
        ReViWo & 2.443 & 1.776 & 1.384 \\
        KYC & 1.653 & 1.143 & 1.361 \\
        \bottomrule
    \end{tabular}
    }
\end{table}

\begin{table}[h]
    \centering
    \caption{\textbf{View and Action Entropy (Stack Three).}
    Entropy-based analysis of representation quality before and after policy fine-tuning in the Stack Three dataset.
    “Seen” and “Unseen” denote view entropies (higher is better, ↑) computed over the 25 views, while “Action” denotes action entropy (lower is better, ↓) based on 10 clustered action classes.
    The upper-bound row corresponds to the entropy of a uniform distribution over 25 views (Seen/Unseen) and 10 action clusters (Action).
    }
    \label{tab:view_action_entropy_stackthree}
    \resizebox{0.85\columnwidth}{!}{
    \begin{tabular}{lrrr}
        \toprule
        \textbf{Model} & \textbf{Seen (↑)} & \textbf{Unseen (↑)} & \textbf{Action (↓)} \\
        \midrule\midrule
        Upper Bound & 3.219 & 3.219 & 2.303 \\
        \midrule\midrule
        \multicolumn{4}{l}{\textbf{Before Fine-tuning}} \\
        \midrule
        \rowcolor{blue!20}
        VILA (Ours) & \textbf{3.111} & \textbf{3.096} & \textbf{0.362} \\
        CLASS & 2.950 & 2.826 & 0.463 \\
        ReViWo & 2.592 & 2.303 & 1.232 \\
        \midrule\midrule
        \multicolumn{4}{l}{\textbf{Fine-Tuned}} \\
        \midrule
        \rowcolor{blue!20}
        VILA (Ours) & \textbf{3.076} & \textbf{3.046} & \textbf{0.339} \\
        Vanilla & 2.986 & 2.883 & 0.488 \\
        CLASS & 2.886 & 2.723 & 0.546 \\
        ReViWo & 2.667 & 2.341 & 1.180 \\
        KYC & 2.118 & 2.073 & 1.053 \\
        \bottomrule
    \end{tabular}
    }
\end{table}

\begin{table}[h]
    \centering
    \caption{\textbf{View and Action Entropy (Coffee).}
    Entropy-based analysis of representation quality before and after policy fine-tuning in the Coffee dataset.
    “Seen” and “Unseen” denote view entropies (higher is better, ↑) computed over the 25 views, while “Action” denotes action entropy (lower is better, ↓) based on 10 clustered action classes.
    The upper-bound row corresponds to the entropy of a uniform distribution over 25 views (Seen/Unseen) and 10 action clusters (Action).
    }
    \label{tab:view_action_entropy_coffee}
    \resizebox{0.85\columnwidth}{!}{
    \begin{tabular}{lrrr}
        \toprule
        \textbf{Model} & \textbf{Seen (↑)} & \textbf{Unseen (↑)} & \textbf{Action (↓)} \\
        \midrule\midrule
        Upper Bound & 3.219 & 3.219 & 2.303 \\
        \midrule\midrule
        \multicolumn{4}{l}{\textbf{Before Fine-tuning}} \\
        \midrule
        \rowcolor{blue!20}
        VILA (Ours) & \textbf{2.554} & \textbf{1.924} & \textbf{0.487} \\
        CLASS & 2.361 & 1.075 & 0.700 \\
        ReViWo & 2.182 & 1.181 & 1.053 \\
        \midrule\midrule
        \multicolumn{4}{l}{\textbf{Fine-Tuned}} \\
        \midrule
        \rowcolor{blue!20}
        VILA (Ours) & \textbf{2.426} & \textbf{1.632} & \textbf{0.488} \\
        Vanilla & 2.204 & 0.809 & 0.710 \\
        CLASS & 0.995 & 0.287 & 0.812 \\
        ReViWo & 2.201 & 1.429 & 1.222 \\
        KYC & 1.274 & 0.457 & 1.010 \\
        \bottomrule
    \end{tabular}
    }
\end{table}

\begin{table}[h]
    \centering
    \caption{\textbf{View and Action Entropy (Mug Cleanup).}
    Entropy-based analysis of representation quality before and after policy fine-tuning in the Mug Cleanup dataset.
    “Seen” and “Unseen” denote view entropies (higher is better, ↑) computed over the 25 views, while “Action” denotes action entropy (lower is better, ↓) based on 10 clustered action classes.
    The upper-bound row corresponds to the entropy of a uniform distribution over 25 views (Seen/Unseen) and 10 action clusters (Action).
    }
    \label{tab:view_action_entropy_mug}
    \resizebox{0.85\columnwidth}{!}{
    \begin{tabular}{lrrr}
        \toprule
        \textbf{Model} & \textbf{Seen (↑)} & \textbf{Unseen (↑)} & \textbf{Action (↓)} \\
        \midrule\midrule
        Upper Bound & 3.219 & 3.219 & 2.303 \\
        \midrule\midrule
        \multicolumn{4}{l}{\textbf{Before Fine-tuning}} \\
        \midrule
        \rowcolor{blue!20}
        VILA (Ours) & \textbf{2.957} & \textbf{2.891} & \textbf{0.452} \\
        CLASS & 2.564 & 2.071 & 0.654 \\
        ReViWo & 2.383 & 1.941 & 1.161 \\
        \midrule\midrule
        \multicolumn{4}{l}{\textbf{Fine-Tuned}} \\
        \midrule
        \rowcolor{blue!20}
        VILA (Ours) & \textbf{2.920} & \textbf{2.793} & \textbf{0.404} \\
        Vanilla & 2.721 & 2.272 & 0.614 \\
        CLASS & 2.153 & 1.569 & 0.824 \\
        ReViWo & 2.437 & 1.904 & 1.302 \\
        KYC & 1.765 & 1.176 & 1.142 \\
        \bottomrule
    \end{tabular}
    }
\end{table}

\subsection{UMAP Visualization}
\begin{figure}
  \centering
  
  \begin{subfigure}{1\columnwidth}
    \includegraphics[width=\linewidth]{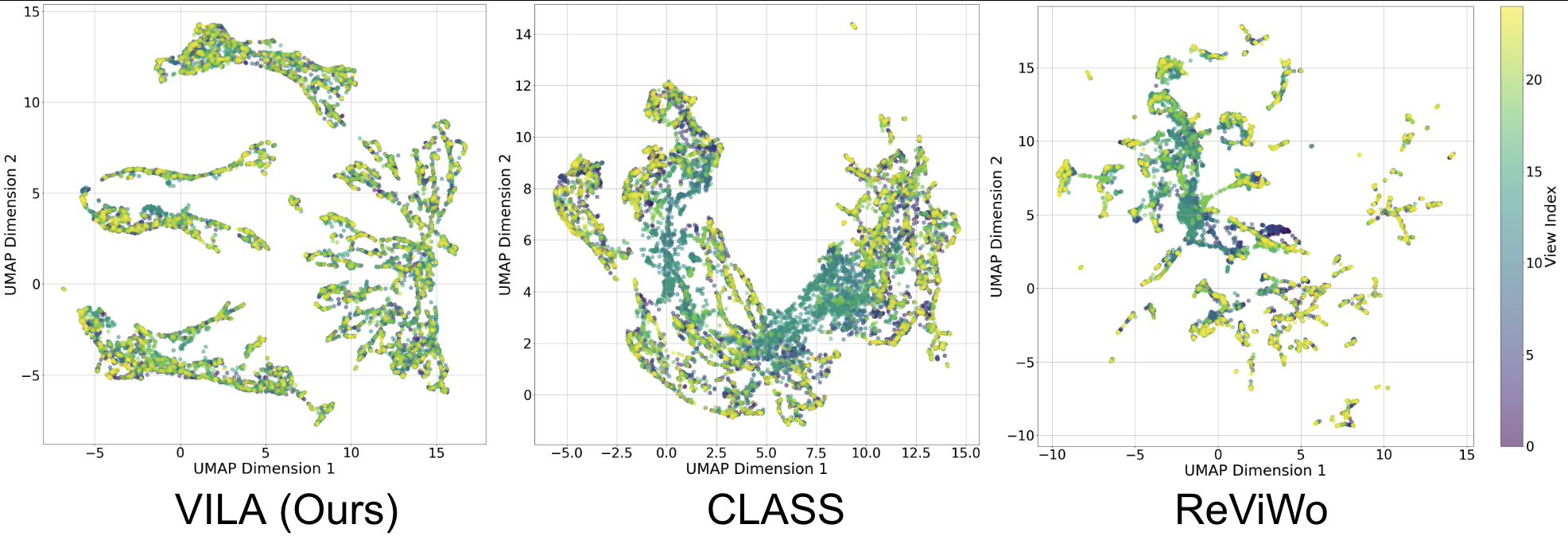}
    \caption{\textbf{Before Policy Training}}
    \label{fig:umap_views_square_before}
  \end{subfigure}
  
  \begin{subfigure}{1\columnwidth}
    \includegraphics[width=\linewidth]{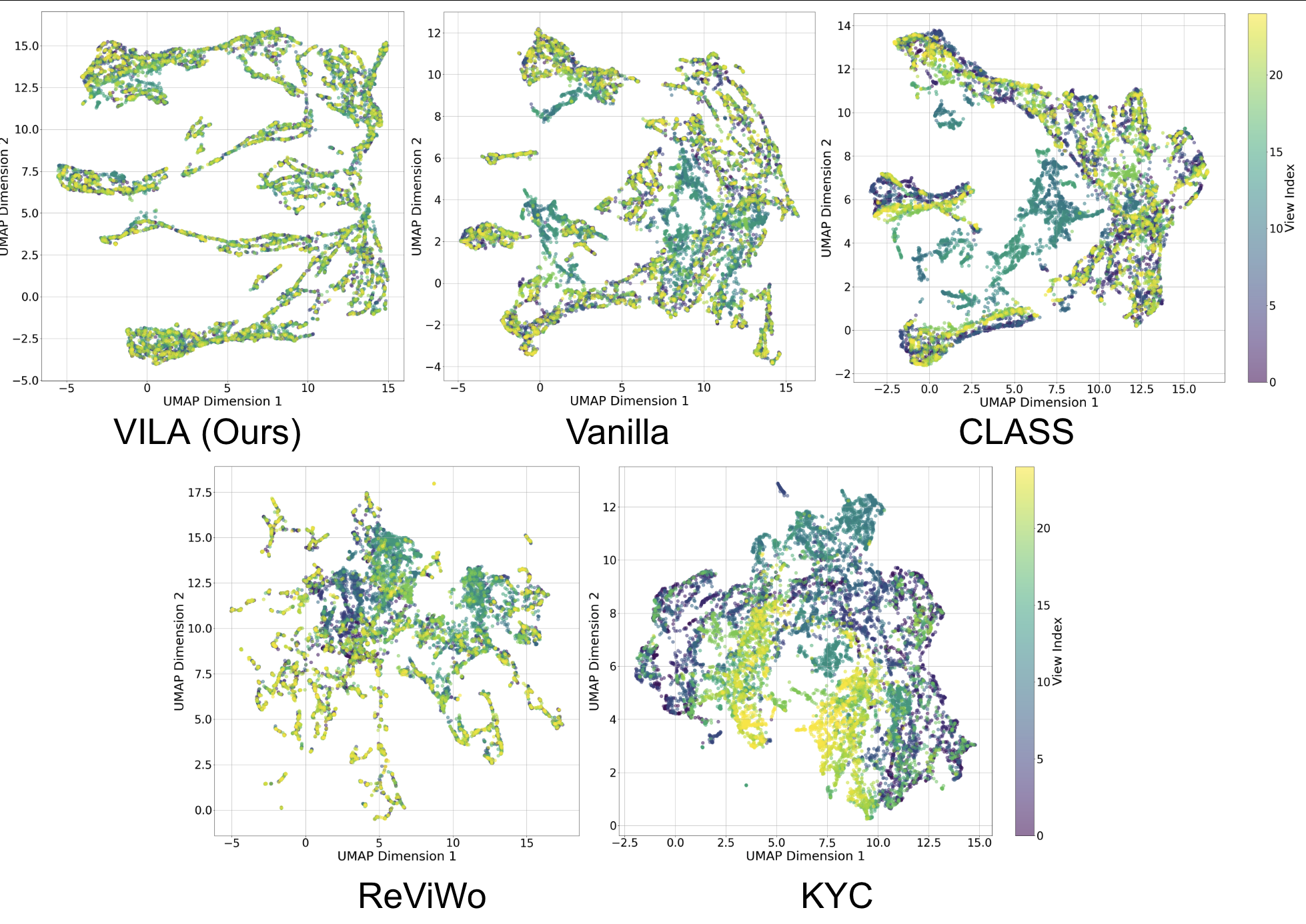}
    \caption{\textbf{After Policy Training}}
    \label{fig:umap_views_square_after}
  \end{subfigure}

  \caption{
    \textbf{UMAP of encoder representations across 25 views on {Square}.}
  }
  \label{fig:umap_square}
\end{figure}

\begin{figure}
  \centering
  
  \begin{subfigure}{1\columnwidth}
    \includegraphics[width=\linewidth]{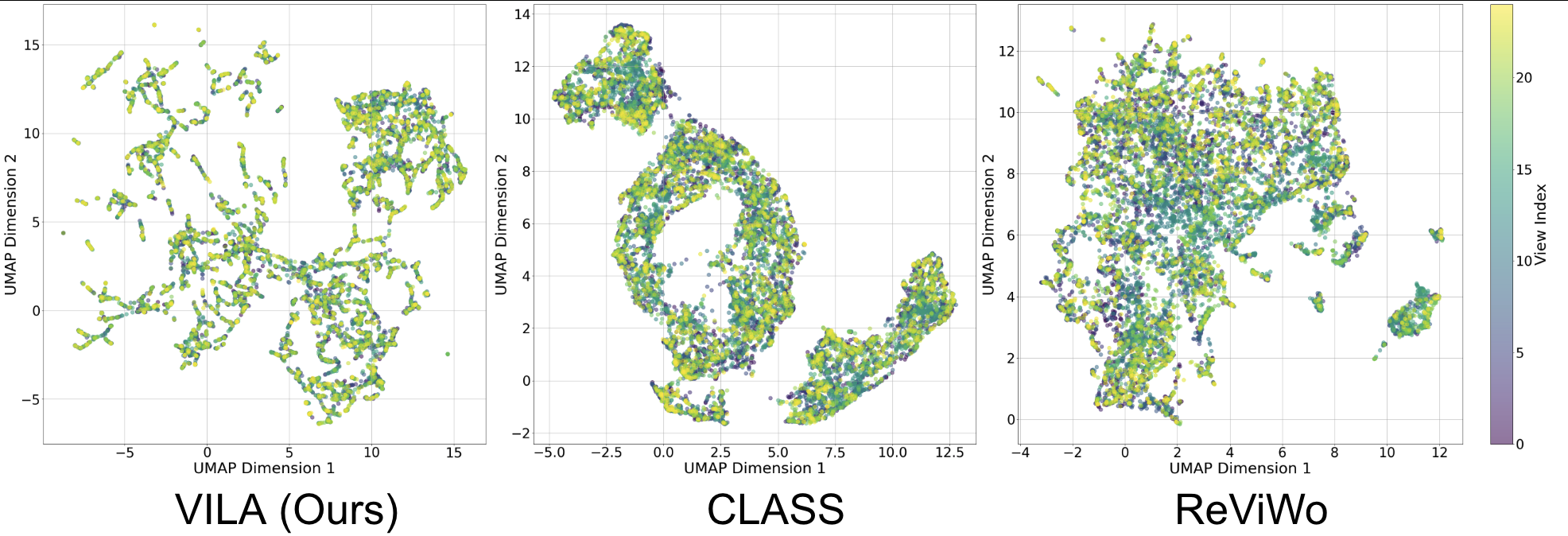}
    \caption{\textbf{Before Policy Training}}
    \label{fig:umap_views_stackthree_before}
  \end{subfigure}
  
  \begin{subfigure}{1\columnwidth}
    \includegraphics[width=\linewidth]{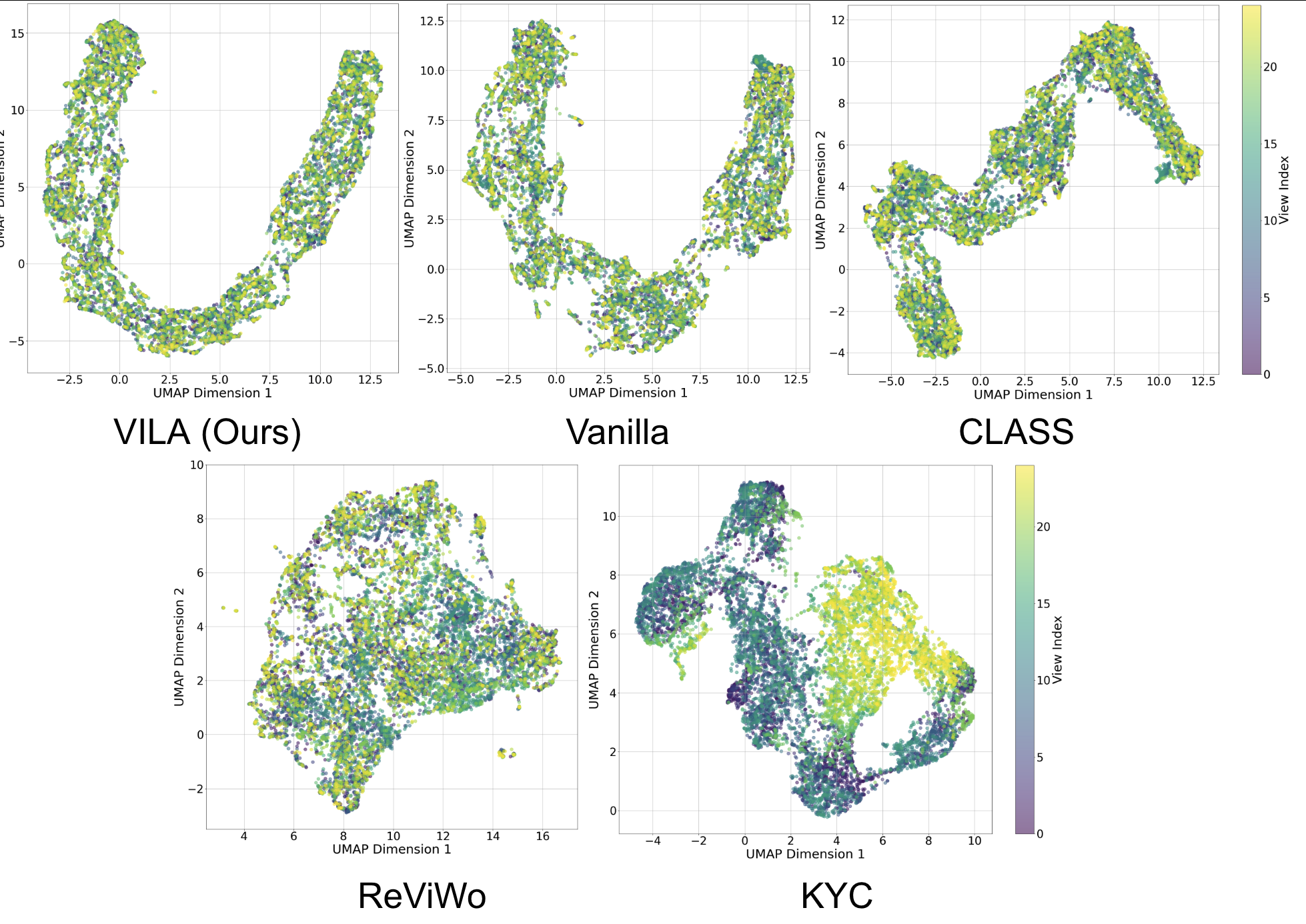}
    \caption{\textbf{After Policy Training}}
    \label{fig:umap_views_stackthree_after}
  \end{subfigure}

  \caption{
    \textbf{UMAP of encoder representations across 25 views on {Stack Three}.}
  }
  \label{fig:umap_stackthree}
\end{figure}

\begin{figure}
  \centering
  
  \begin{subfigure}{1\columnwidth}
    \includegraphics[width=\linewidth]{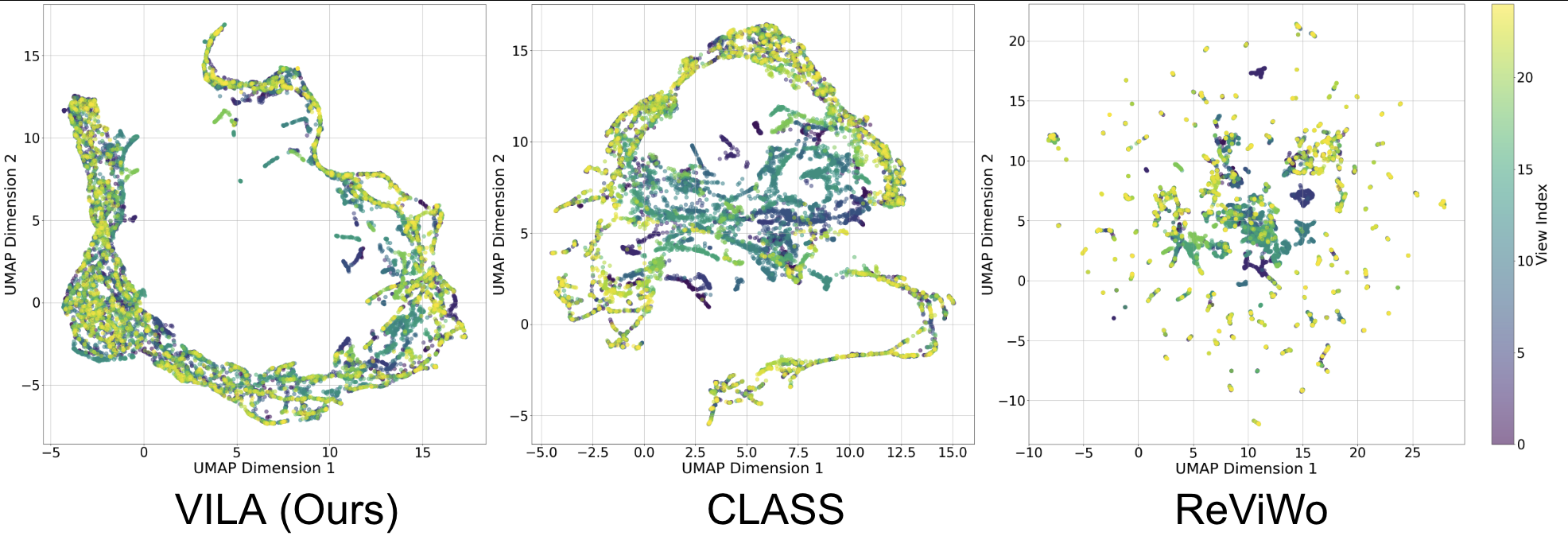}
    \caption{\textbf{Before Policy Training}}
    \label{fig:umap_views_coffee_before}
  \end{subfigure}
  
  \begin{subfigure}{1\columnwidth}
    \includegraphics[width=\linewidth]{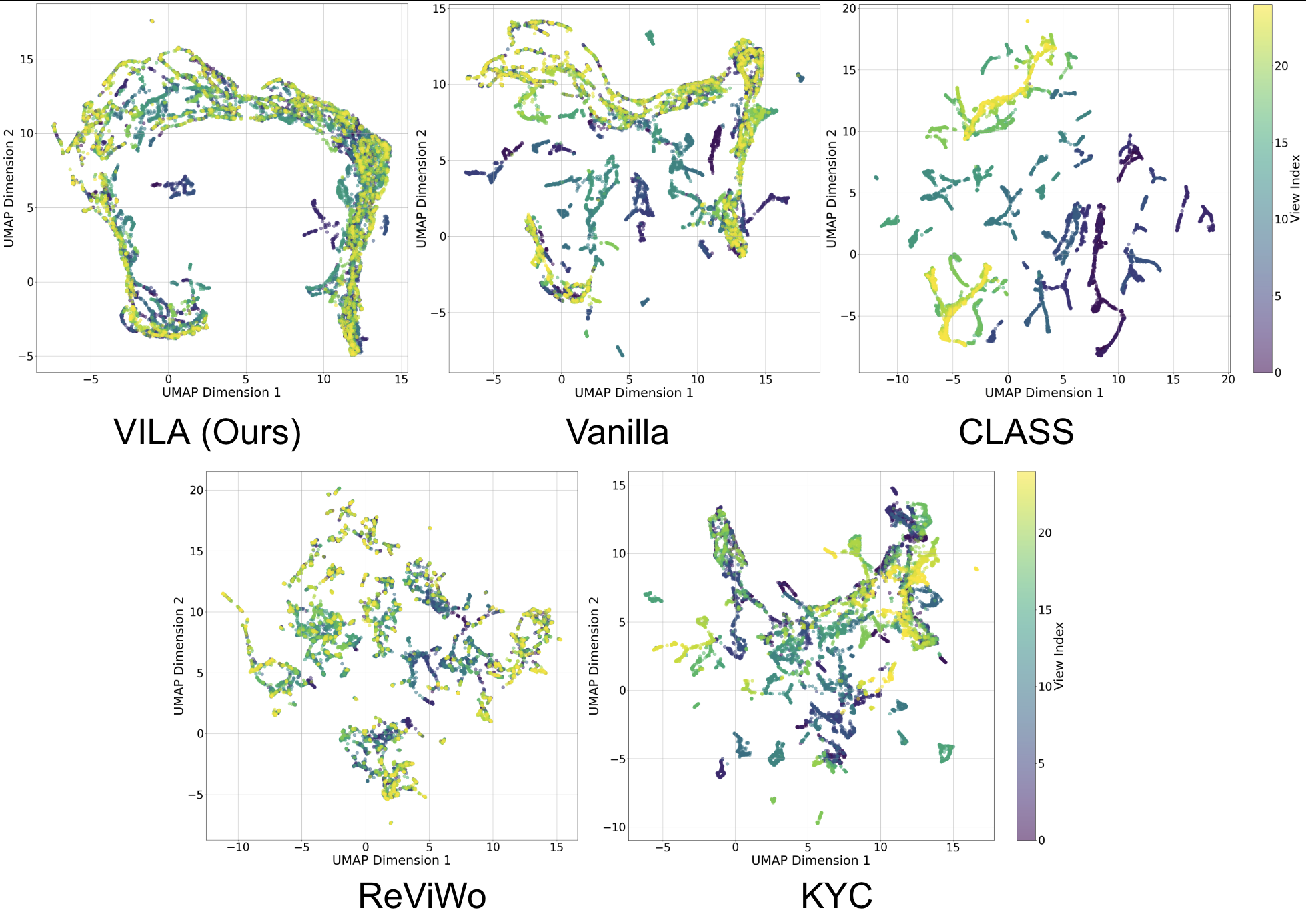}
    \caption{\textbf{After Policy Training}}
    \label{fig:umap_views_coffee_after}
  \end{subfigure}

  \caption{
    \textbf{UMAP of encoder representations across 25 views on {Coffee}.}
  }
  \label{fig:umap_coffee}
\end{figure}

\begin{figure}
  \centering
  
  \begin{subfigure}{1\columnwidth}
    \includegraphics[width=\linewidth]{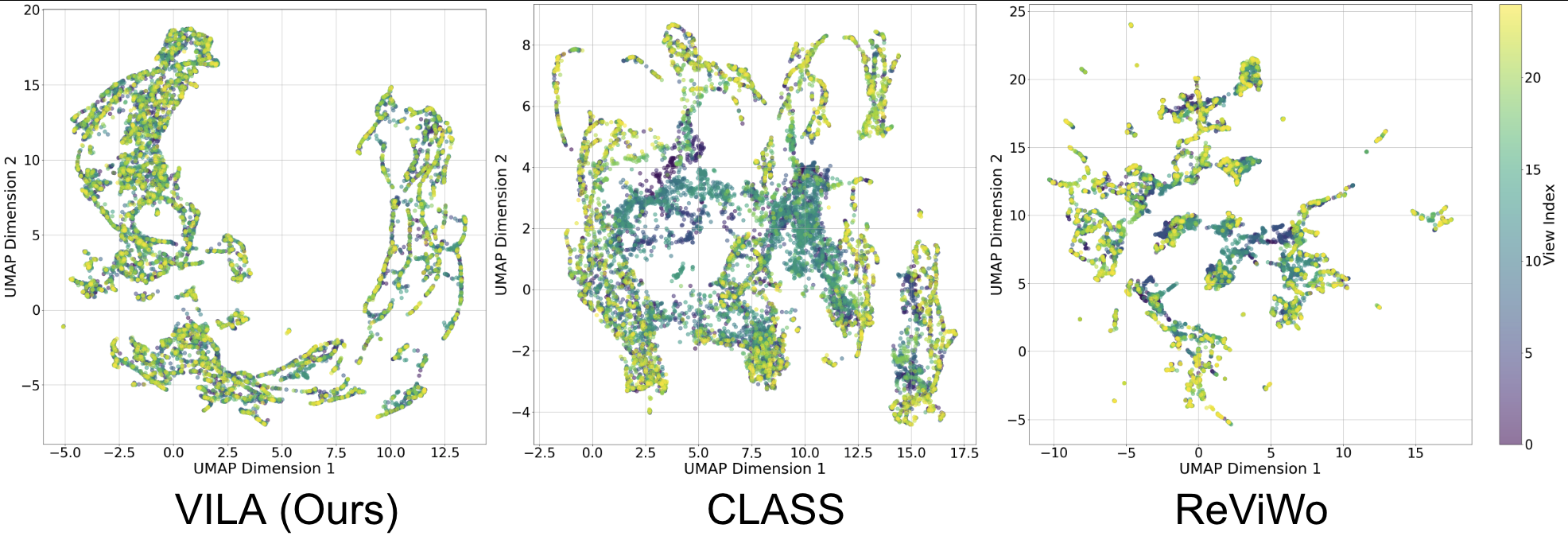}
    \caption{\textbf{Before Policy Training}}
    \label{fig:umap_views_mugcleanup_before}
  \end{subfigure}
  
  \begin{subfigure}{1\columnwidth}
    \includegraphics[width=\linewidth]{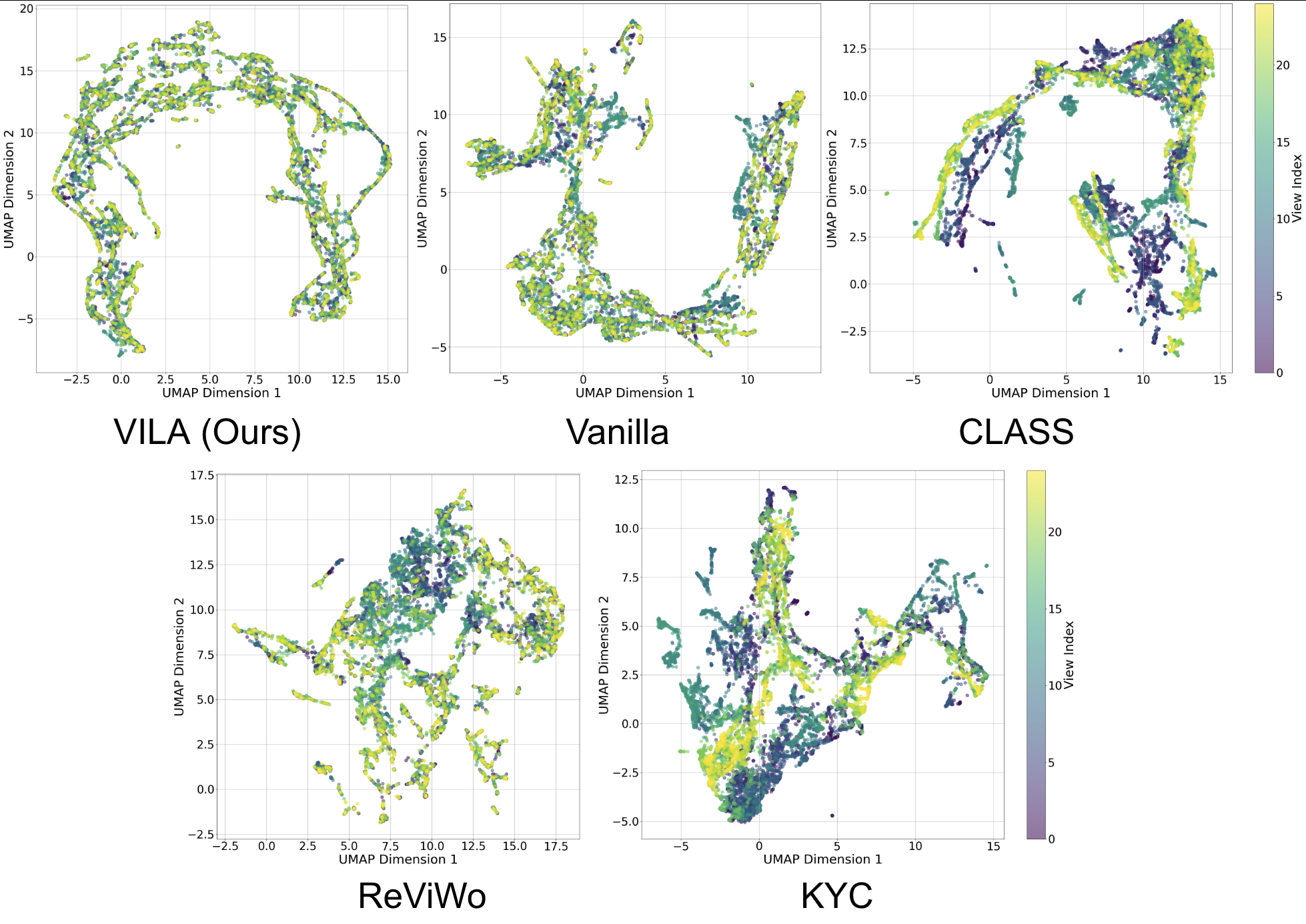}
    \caption{\textbf{After Policy Training}}
    \label{fig:umap_views_mugcleanup_after}
  \end{subfigure}

  \caption{
    \textbf{UMAP of encoder representations across 25 views on {Mug Cleanup}.}
  }
  \label{fig:umap_mugcleanup}
\end{figure}

\paragraph{Additional view-based UMAPs.}
In Figures~\ref{fig:umap_square}--\ref{fig:umap_mugcleanup}, 
we additionally visualize encoder representations across the 25 camera views on 
{Square}, {Stack-Three}, {Coffee}, and {Mug-Cleanup}. 

\begin{figure}
  \centering
  
  \begin{subfigure}{1\columnwidth}
    \includegraphics[width=\linewidth]{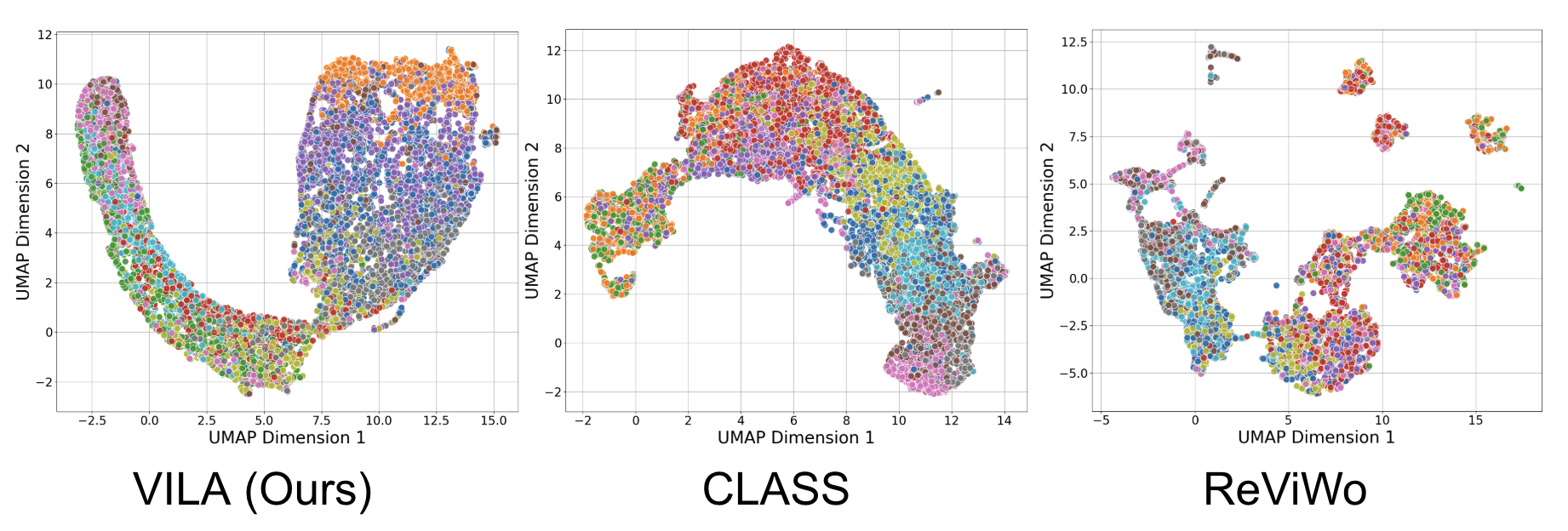}
    \caption{\textbf{Before Policy Training}}
    \label{fig:umap_act_lift_before}
  \end{subfigure}
  
  \begin{subfigure}{1\columnwidth}
    \includegraphics[width=\linewidth]{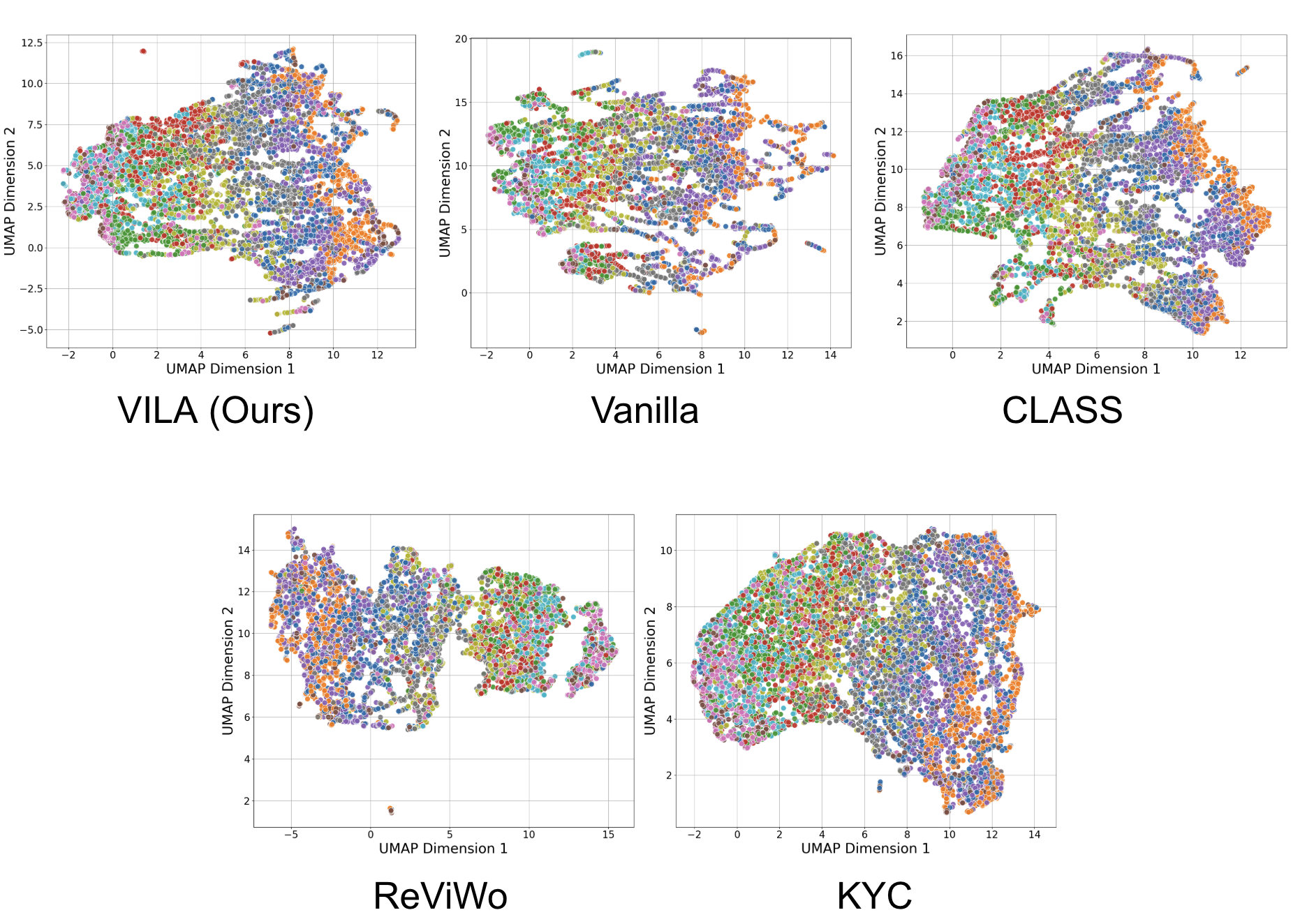}
    \caption{\textbf{After Policy Training}}
    \label{fig:umap_act_lift_after}
  \end{subfigure}

  \caption{
    \textbf{UMAP of encoder representations colored by action clusters on {Lift}.}
  }
  \label{fig:umap_act_lift}
\end{figure}

\begin{figure}
  \centering
  
  \begin{subfigure}{1\columnwidth}
    \includegraphics[width=\linewidth]{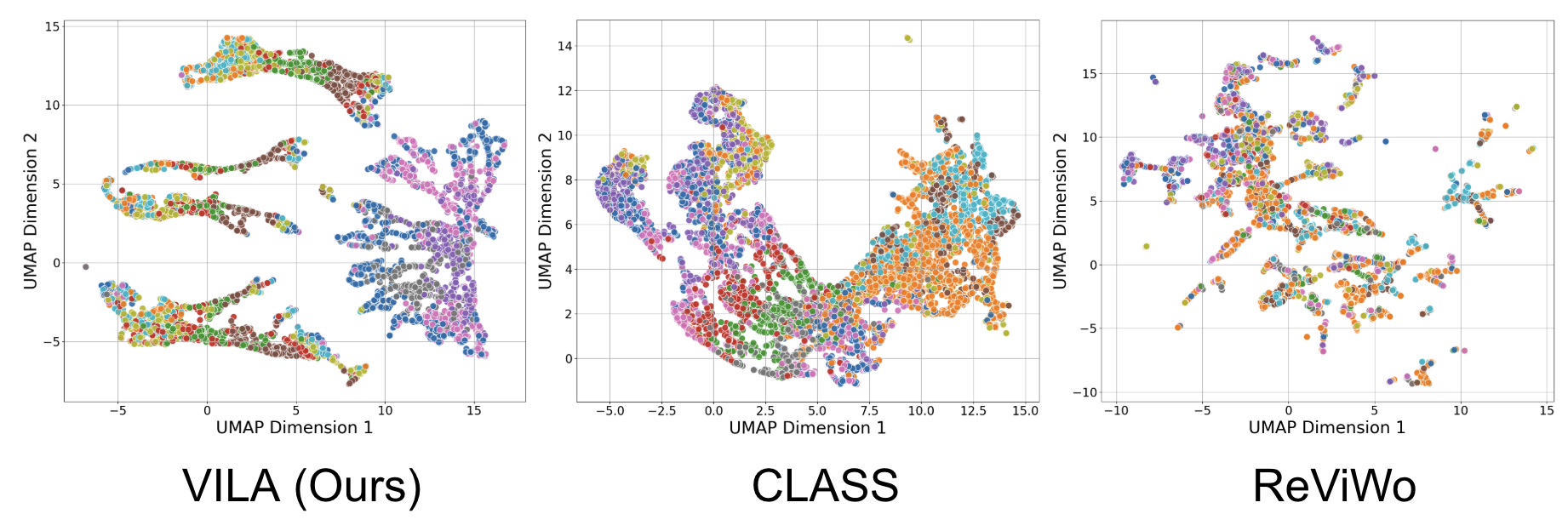}
    \caption{\textbf{Before Policy Training}}
    \label{fig:umap_act_square_before}
  \end{subfigure}
  
  \begin{subfigure}{1\columnwidth}
    \includegraphics[width=\linewidth]{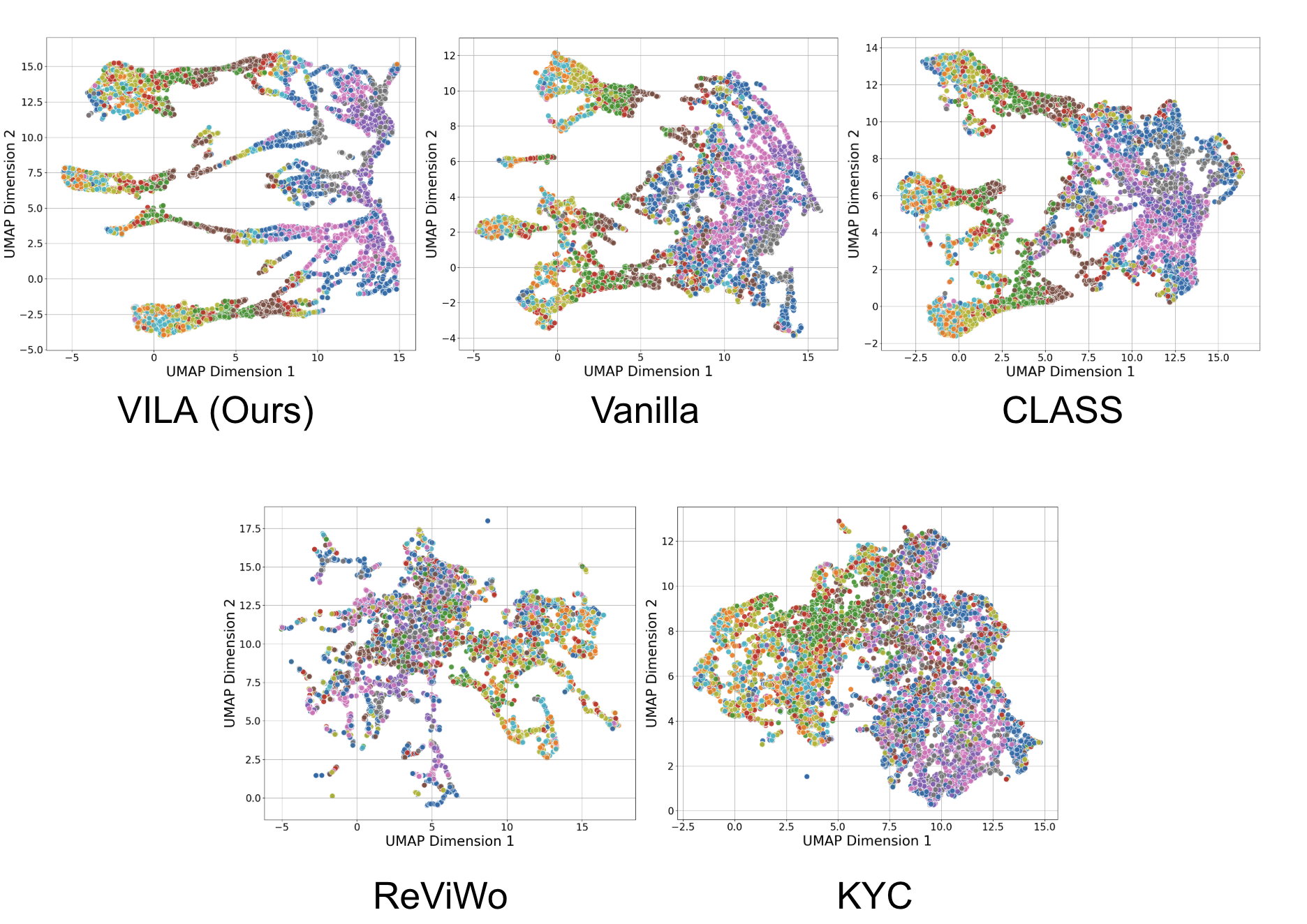}
    \caption{\textbf{After Policy Training}}
    \label{fig:umap_act_square_after}
  \end{subfigure}

  \caption{
    \textbf{UMAP of encoder representations colored by action clusters on {Square}.}
  }
  \label{fig:umap_act_square}
\end{figure}

\begin{figure}
  \centering
  
  \begin{subfigure}{1\columnwidth}
    \includegraphics[width=\linewidth]{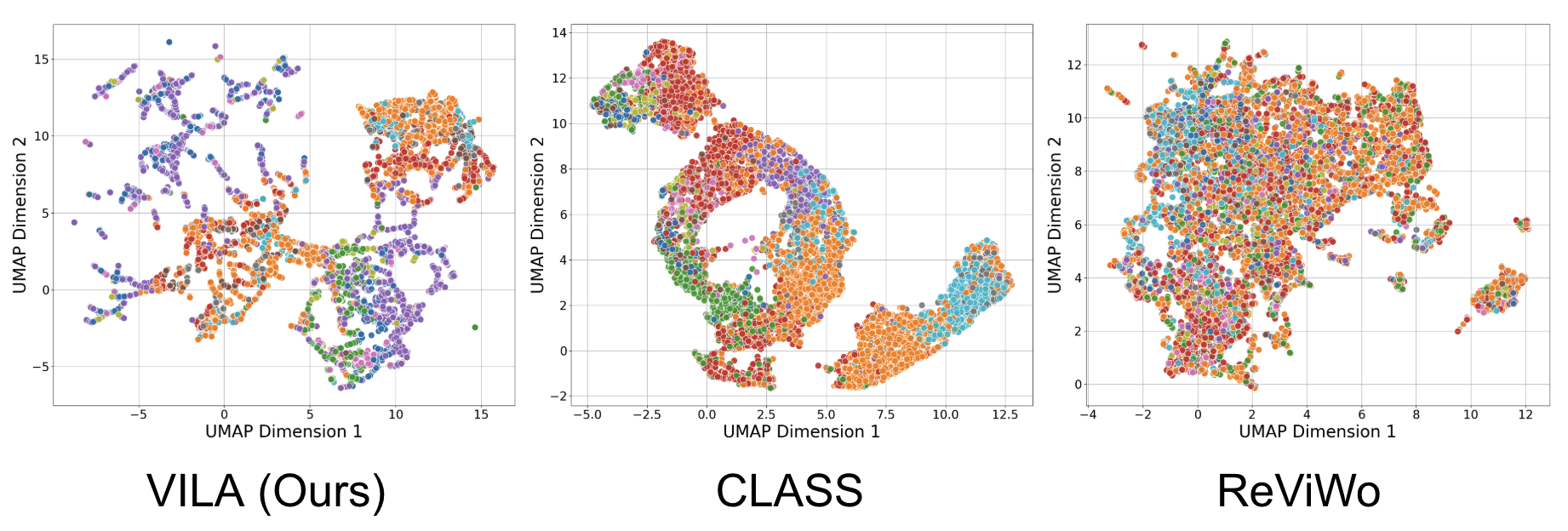}
    \caption{\textbf{Before Policy Training}}
    \label{fig:umap_act_stackthree_before}
  \end{subfigure}
  
  \begin{subfigure}{1\columnwidth}
    \includegraphics[width=\linewidth]{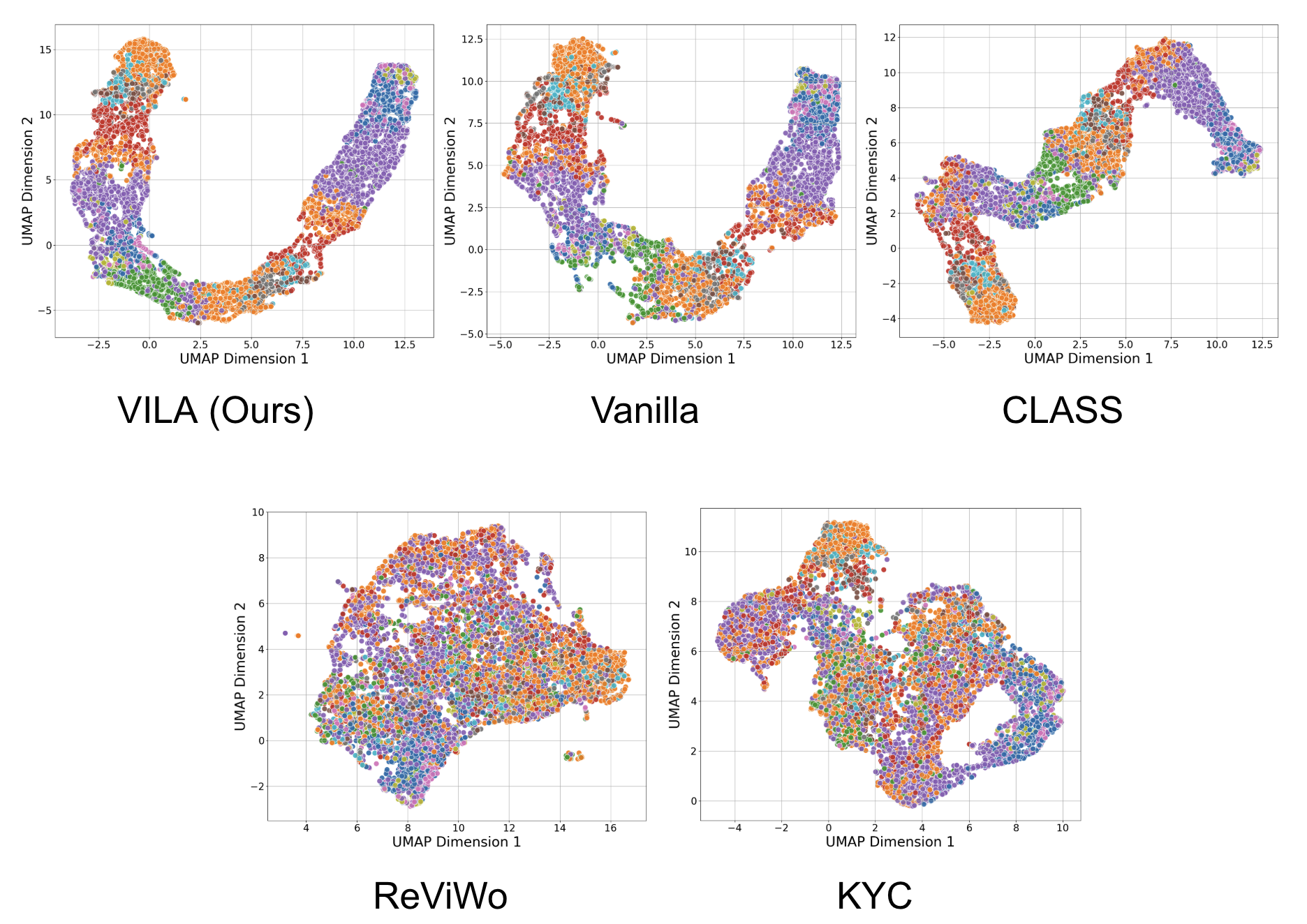}
    \caption{\textbf{After Policy Training}}
    \label{fig:umap_act_stackthree_after}
  \end{subfigure}

  \caption{
    \textbf{UMAP of encoder representations colored by action clusters on {Stack Three}.} 
  }
  \label{fig:umap_act_stackthree}
\end{figure}

\begin{figure}
  \centering
  
  \begin{subfigure}{1\columnwidth}
    \includegraphics[width=\linewidth]{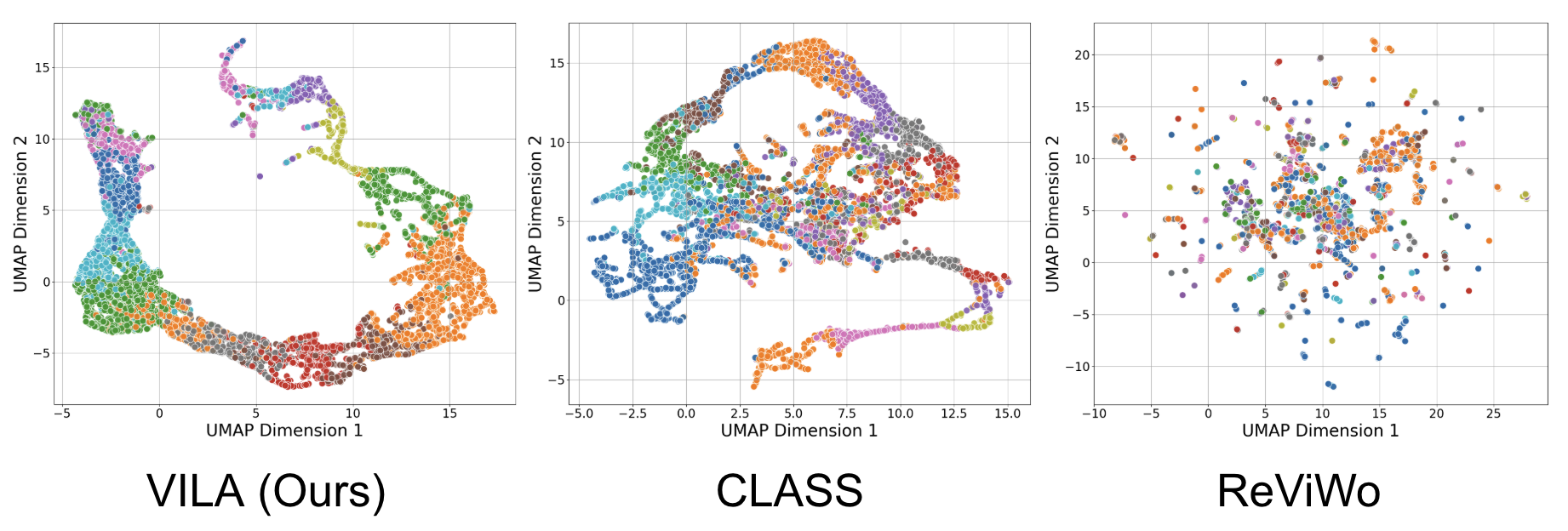}
    \caption{\textbf{Before Policy Training}}
    \label{fig:umap_act_coffee_before}
  \end{subfigure}
  
  \begin{subfigure}{1\columnwidth}
    \includegraphics[width=\linewidth]{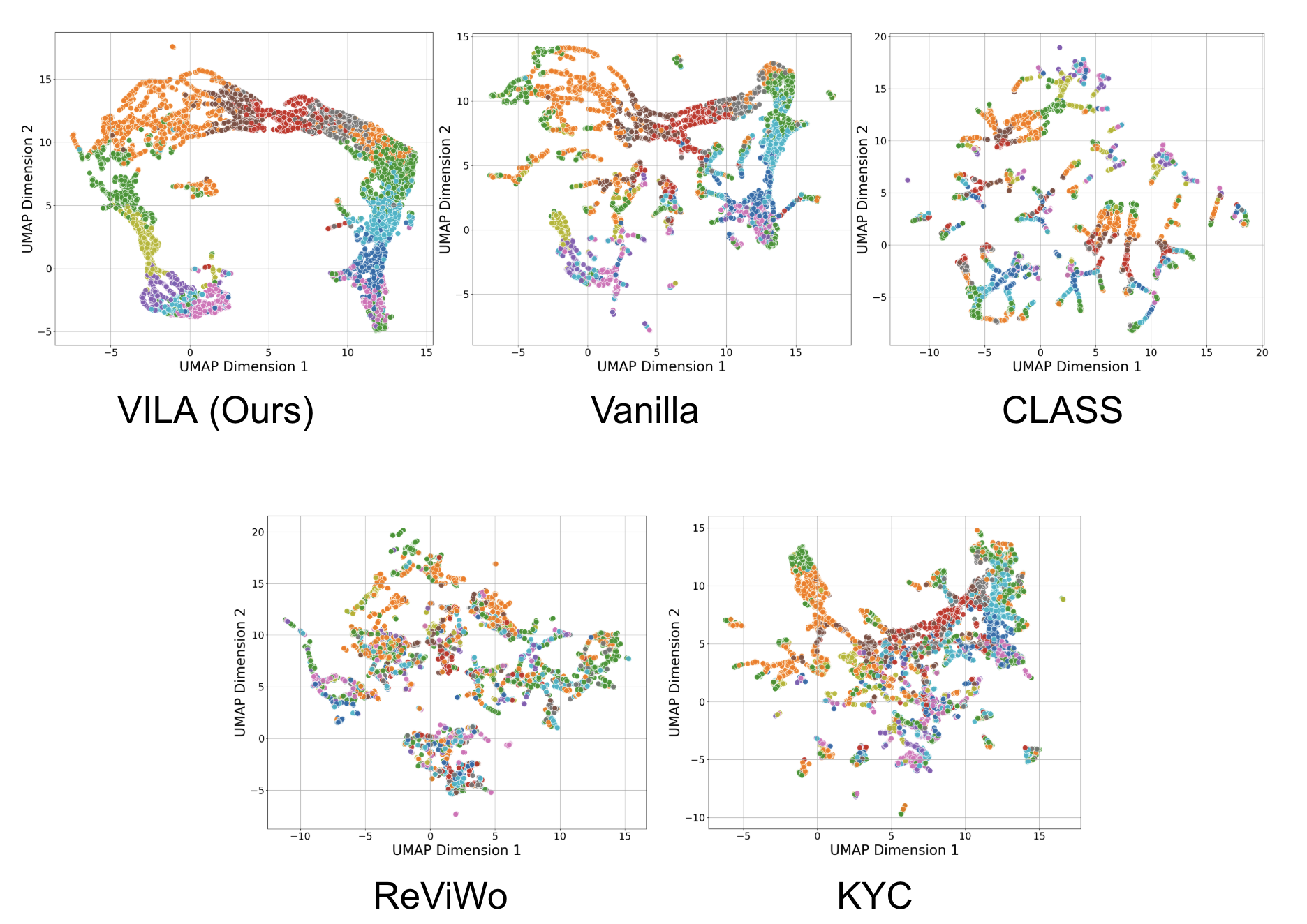}
    \caption{\textbf{After Policy Training}}
    \label{fig:umap_act_coffee_after}
  \end{subfigure}

  \caption{
    \textbf{UMAP of encoder representations colored by action clusters on {Coffee}.}
  }
  \label{fig:umap_act_coffee}
\end{figure}

\begin{figure}
  \centering
  
  \begin{subfigure}{1\columnwidth}
    \includegraphics[width=\linewidth]{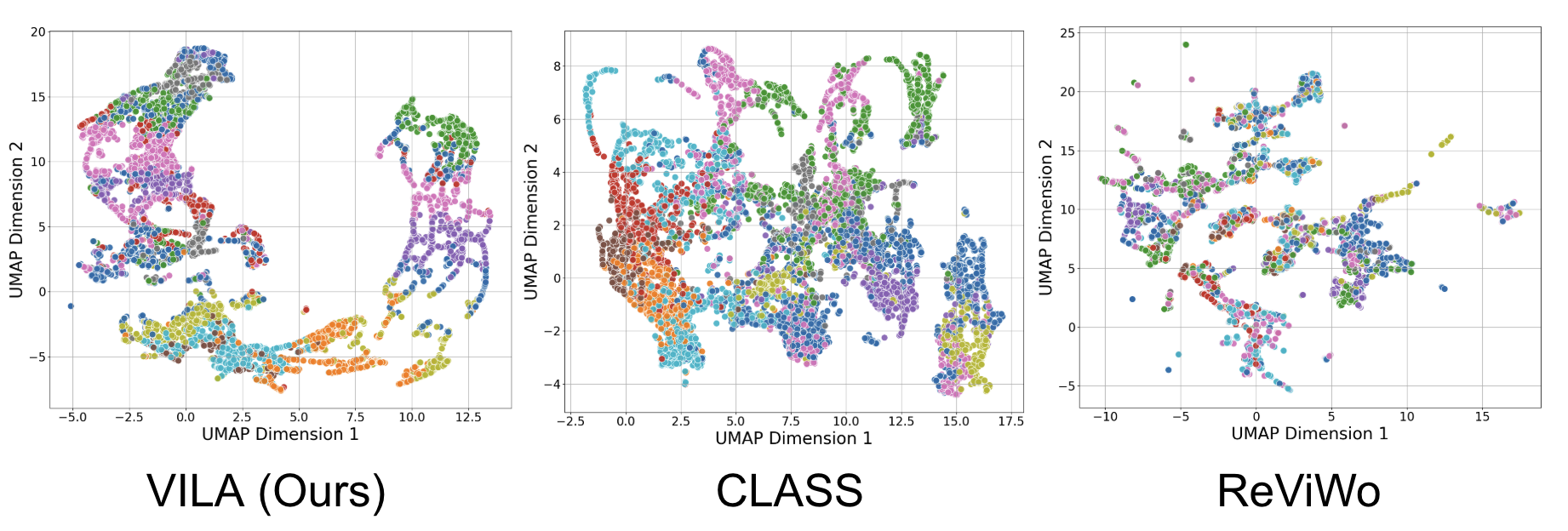}
    \caption{\textbf{Before Policy Training}}
    \label{fig:umap_act_mugcleanup_before}
  \end{subfigure}
  
  \begin{subfigure}{1\columnwidth}
    \includegraphics[width=\linewidth]{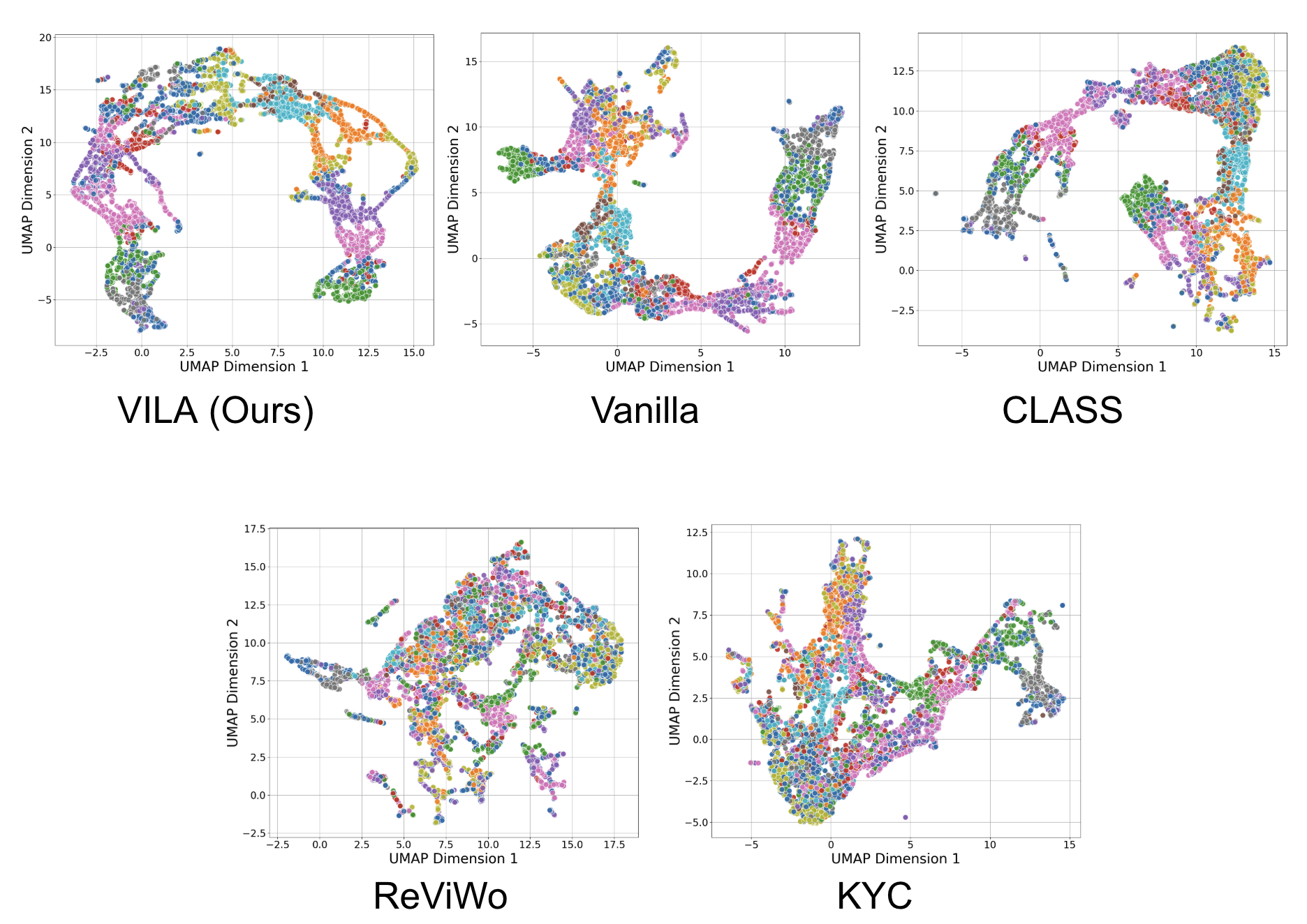}
    \caption{\textbf{After Policy Training}}
    \label{fig:umap_act_mugcleanup_after}
  \end{subfigure}

  \caption{
    \textbf{UMAP of encoder representations colored by action clusters on {Mug Cleanup}.}
  }
  \label{fig:umap_act_mugcleanup}
\end{figure}

\paragraph{Action-cluster UMAPs.}
In Figures~\ref{fig:umap_act_lift}--\ref{fig:umap_act_mugcleanup}, 
we reuse the same encoder representations as in the view-based UMAPs, 
but color them by $K{=}10$ action clusters obtained by applying $k$-means to 
10-step GT action sequences $\{a_t,\dots,a_{t+9}\}$. 



\end{document}